\pdfoutput=1
\documentclass[11pt]{article}

\usepackage[preprint]{acl}
\usepackage{times}
\usepackage{latexsym}

\usepackage[T1]{fontenc}
\usepackage[utf8]{inputenc}
\usepackage{amssymb}
\usepackage{pifont}
\usepackage{microtype}
\usepackage{inconsolata}
\usepackage[most]{tcolorbox}
\usepackage{graphicx}
\usepackage{array}
\usepackage{hyperref}

\usepackage{tikz}
\usepackage{tabularx}
\usepackage{subfigure,graphicx}
\usepackage{booktabs}
\usepackage{multirow}
\usepackage{makecell}
\usepackage{xcolor}
\usepackage{eqparbox} 
\usepackage{transparent}

\newcommand{\ie}{\textit{i.e.}}
\newcommand{\eg}{\textit{e.g.}}

\newcommand{\ours}{{OmniEval}}
\newcommand{\TM}{{T$^2$M}}

\title{\includegraphics[width=0.04\textwidth]{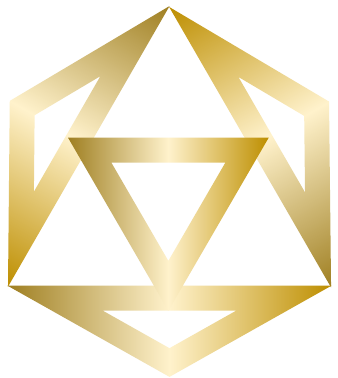}\;
\ours{}: An Omnidirectional and Automatic RAG Evaluation Benchmark in Financial Domain}

\author{Shuting Wang$^{1,\dagger}$, Jiejun Tan$^{1,\dagger}$, \textbf{Zhicheng Dou$^{1*}$}, \and \textbf{Ji-Rong Wen$^{1}$}  \\
$^1$Gaoling School of Artificial Intelligence, Renmin University of China \\
\texttt{\{wangshuting, zstanjj, dou\}@ruc.edu.cn}
}

\begin{document}

\maketitle
\def\thefootnote{*}\footnotetext{Corresponding author.}
\def\thefootnote{$\dagger$}\footnotetext{Equal contribution.} 
\begin{abstract}

Retrieval-augmented generation (RAG) has emerged as a key application of large language models (LLMs), especially in vertical domains where LLMs may lack domain-specific knowledge. This paper introduces \ours{}, an omnidirectional and automatic RAG benchmark for the financial domain, featured by its multi-dimensional evaluation framework: 
First, we categorize RAG scenarios by five task classes and 16 financial topics, leading to a \textit{matrix-based structured assessment} for RAG evaluation; 
Next, we leverage a \textit{multi-dimensional evaluation data generation method} that integrates GPT-4-based automatic generation and human annotation approaches, achieving an 87.47\% acceptance ratio in human evaluations of generated instances; 
Further, we utilize a \textit{multi-stage evaluation pipeline} to assess both retrieval and generation performance, resulting in an all-sided evaluation of the RAG pipeline.  
Finally, rule-based and LLM-based metrics are combined to build a \textit{multi-dimensional evaluation system}, enhancing the reliability of assessments through fine-tuned LLM-based evaluators. 
Our omnidirectional evaluation experiments highlight the performance variations of RAG systems across diverse topics and tasks and reveal significant opportunities for RAG models to improve their capabilities in vertical domains. 
The code link of our benchmark is \href{https://github.com/RUC-NLPIR/OmniEval}{https://github.com/RUC-NLPIR/OmniEval}
\end{abstract}

\section{Introduction}
RAG techniques have gained prominence as one of the most widespread and practical applications of LLMs. Particularly in specialized domains where LLMs often lack in-domain expertise, RAG models effectively incorporate external domain corpora and the internal knowledge of LLMs to enhance the overall quality of generative AI systems. 
Despite the advancements, the challenge of automatically building high-quality omnidirectional benchmarks to evaluate the performance of RAG models within specific vertical domains remains unresolved. In this study, we introduce an automatic and omnidirectional benchmark, \ours{}, designed to assess RAG systems in a widely adopted vertical domain, finance. Our proposed benchmark illustrates its versatility and automaticity from the following angles: 

\textit{Matrix-based RAG scenario evaluation.} 
Versatile response capabilities are essential for RAG systems to handle diverse user queries spanning various scenarios. 
For example, some queries seek factual information that can be extracted from web pages, while others may require complex financial computations. To evaluate such versatility, we classified RAG scenarios into five common tasks, \ie, extractive question-answering (QA), multi-hop reasoning, contrast QA, long-form QA, and conversational QA. Moreover, in specialized domains like finance, user queries often fall into distinct domain topics. Consequently, we also distinguish RAG scenarios based on topical categories of queries, recognizing 16 common subcategories in the finance domain. 
These two orthogonal taxonomies lead to matrix-based RAG evaluation scenarios and support all-sided profiles for RAG systems. 

\textit{Multi-dimensional evaluation data generation.} To create extensible and high-quality evaluation datasets, we integrate the GPT-4-based automated generation and human annotation approaches. The former provides flexibility, allowing the data generation pipeline to adapt to various domains, and the latter guarantees the quality of the datasets. Our human evaluation of automated generated instances indicates an acceptable ratio of 87.47\%, confirming the effectiveness of our data generation pipeline. 

\textit{Multi-stage evaluation.} The quality of the retrieval and generation processes are both important when evaluating the RAG pipeline, especially for vertical domains, since general retrievers may lack expert knowledge and potentially compromise the response quality. Therefore, \ours{} evaluates both retriever and generator performance to provide a comprehensive assessment for RAG systems. 

\textit{Multi-dimensional evaluation metrics.} For the evaluation systems, we build our evaluation metrics by combining rule-based and LLM-based metrics together. The former embodies widely used evaluation metrics, such as MAP and Rouge, offering solid evaluation results. The latter is produced from fine-tuned LLMs to achieve high-level evaluation beyond term-level matching, such as hallucination detection and numerical accuracy. To ensure the reliability of our LLM-based evaluation, we further manually annotate some evaluation samples and fine-tune Qwen2.5-7B-Instruct~\cite{qwen2.5} to build LLM evaluators.

As a result, \ours{} contains 11.4k automatically generated test examples and 1.7k human-annotated test examples. We further split out 3k automatically generated examples as a training set for future investigations.\footnote{Note that the automatically generated examples are extensible by prompting GPT-4~\cite{GPT4}, we currently provide this amount of examples due to the limited budgets.} 
The preliminary assessment of our LLM evaluators indicates that they significantly surpass prompting-based LLMs in evaluation abilities, demonstrating 74.4\% accuracy. 

Our evaluation experiments are conducted on various retrievers, including BGE-M3~\cite{bge-m3}, BGE-large-zh~\cite{bge-large-zh}, GTE-Qwen2-1.5b~\cite{gte}, and jina-zh~\cite{jina-zh}, and diverse open-resource LLMs, \ie, Qwen2.5-72B-Instruct~\cite{qwen2.5}, Llama3.1-70B-Instruct~\cite{llama31}, Deepseek-v2-chat~\cite{deepseekv2}, and Yi15-34B~\cite{yi}. The experimental results reveal that RAG performance varies across different topics and tasks. Moreover, there remains a large space to improve RAG systems in vertical domains. 

\begin{table*}[!ht]
    \centering
    \resizebox{0.99\linewidth}{!}{\begin{tabular}{lccccccccc}
    \toprule
    \multirow{2}{*}{Benchmark} & \multicolumn{2}{c}{Evaluation Scenarios} & \multicolumn{2}{c}{Data Generation}     & \multicolumn{3}{c}{Evaluation Metrics} & \multicolumn{2}{c}{Evaluation   Models} \\
    \cmidrule(lr){2-3}\cmidrule(lr){4-5}\cmidrule(lr){6-8}\cmidrule(lr){9-10}
    & Task-Spe. & Topic-Spe. & Manual & Auto. & Rule & Model & Human & Retriever & Generator \\
    \midrule
    PIXIU~\cite{PIXIU}  &  $\checkmark$ & \ding{55} & \ding{55} & \ding{55} & $\checkmark$ & \ding{55} & $\checkmark$ & \ding{55} & $\checkmark$  \\
    DISC-FinLLM~\cite{DISC-FinLLM} & $\checkmark$ & \ding{55} & \ding{55} & $\checkmark$ & $\checkmark$ & $\checkmark$ & \ding{55} & \ding{55} & $\checkmark$  \\
    FinanceBench~\cite{FinanceBench} & $\checkmark$ & $\checkmark$ & $\checkmark$ & \ding{55} & \ding{55} & \ding{55} & $\checkmark$ & \ding{55} & $\checkmark$  \\
    AlphaFin~\cite{AlphaFin} &  $\checkmark$ & \ding{55} & \ding{55} & \ding{55} & $\checkmark$ & $\checkmark$ & $\checkmark$ & \ding{55} & $\checkmark$  \\
    FinBen~\cite{FinBen} & $\checkmark$ & \ding{55} & \ding{55} & \ding{55} & $\checkmark$ & $\checkmark$ & \ding{55} & \ding{55} & $\checkmark$ \\
    FinTextQA~\cite{FinTextQA} & $\checkmark$ & \ding{55} & \ding{55} & \ding{55} & $\checkmark$ & $\checkmark$ & \ding{55} & $\checkmark$ & $\checkmark$  \\
    \midrule
    \ours{} & $\checkmark$ & $\checkmark$ & $\checkmark$ & $\checkmark$ & $\checkmark$ & $\checkmark$ & $\checkmark$ & $\checkmark$ & $\checkmark$  \\
    \bottomrule
    \end{tabular}}
    \caption{The comparison between our proposed benchmark and existing financial benchmarks. ``Auto.'' is short for ``Auto-generated'', ``Spe.'' is short for ``Specific''.}
    \label{tab:bench_compair}
\end{table*}
\begin{table}[!h]
% \small
    \centering
    \resizebox{0.99\linewidth}{!}{
    \begin{tabular}{llrr}
    \toprule
    Datasource	& Data Type & Doc Number & Length Sum  \\
    \midrule
    BSCF-DB & DB - JSON & 193,774 & 23,631,875  \\
    BSCF-PDF & PDF - TXT & 3,082 & 10,587,648  \\
    FinGLM & PDF - TXT & 55,595 & 97,296,690  \\
    Wiki-Fin & JSON & 3,367 & 5,679,758  \\
    BAAI-Fin & JSON & 48,124 & 70,014,858  \\
    Official Web & JSON & 58,616 & 45,837,298  \\
    \bottomrule
    \end{tabular}
    }
    \caption{Statistical information of our diverse data sources. ``Doc'' and ``Sum'' are short for ``Document'' and ``Summation''.}
    \label{tab:statistic_data_source}
\end{table}
\section{Related Work}
\label{sec:related_work}
\subsection{RAG Benchmarks}
With the rapid development of RAG investigation, existing QA datasets and evaluation metrics are limited to providing advanced evaluation results. Therefore, various researchers~\cite{RGB,RECALL,Bench-Med,ARES,ReEval,CRUD-RAG,DomainRAG} concentrate on building comprehensive and reliable RAG benchmarks. The early study, RGB~\cite{RGB}, focuses on the advanced abilities of RAG models, such as noise robustness and information integration. ARES~\cite{ARES} automatically builds a RAG benchmark with the support of LLMs, including automatically generating data instances and automatically judging responses. Beyond open-domain QA, some studies~\cite{Bench-Med, DomainRAG} also constructed domain-specific RAG benchmarks to evaluate the abilities of RAG systems in vertical domains.

\subsection{LLM Evaluation in Financial Domains}
In practice, finance is one of the most widespread vertical domains, comprising a wealth of professional knowledge. Therefore, evaluating LLMs in the financial domain is critical for assessing their expertise in vertical domains. 
Some studies~\cite{FLUE,PIXIU,FinBen,AlphaFin,DISC-FinLLM} collect existing financial QA datasets~\cite{beir,SA-News,salinas-alvarado-etal-2015-domain,chen2021finqa,chen2022convfinqa,soun2022accurate} to build benchmarks, thereby assessing LLMs' understanding of financial knowledge. 
Recently, \citet{PIXIU} further develops instruction-tuning financial benchmarks by writing instructions for various financial tasks. Beyond assessing LLMs alone, AlphaFin~\cite{AlphaFin} also introduces RAG tasks to judge RAG models on financial scenarios. However, it primarily focuses on the quality of final responses, neglecting the retrieval performance. In this paper, we construct an omnidirectional and automatic RAG evaluation benchmark that automatically generates evaluation datasets and omnidirectionally assesses RAG systems, leading to comprehensive profiles for them. 
We compare our benchmark to existing financial LLM benchmarks in Table~\ref{tab:bench_compair} to demonstrate our advantages.

\section{Construction Pipeline of \ours{}}
We introduce the construction pipeline of our benchmark alongside the following steps: 
First, we demonstrate the collection of knowledge corpus in Section~\ref{sec:knowledgecorpus}. Next, the generation of evaluation instances is illustrated in Section~\ref{sec:datagen}. Finally, in Section~\ref{sec:evaluation}, we introduce the evaluation of RAG models. The details are demonstrated below.

\begin{figure}
    \centering
    \includegraphics[width=1\linewidth]{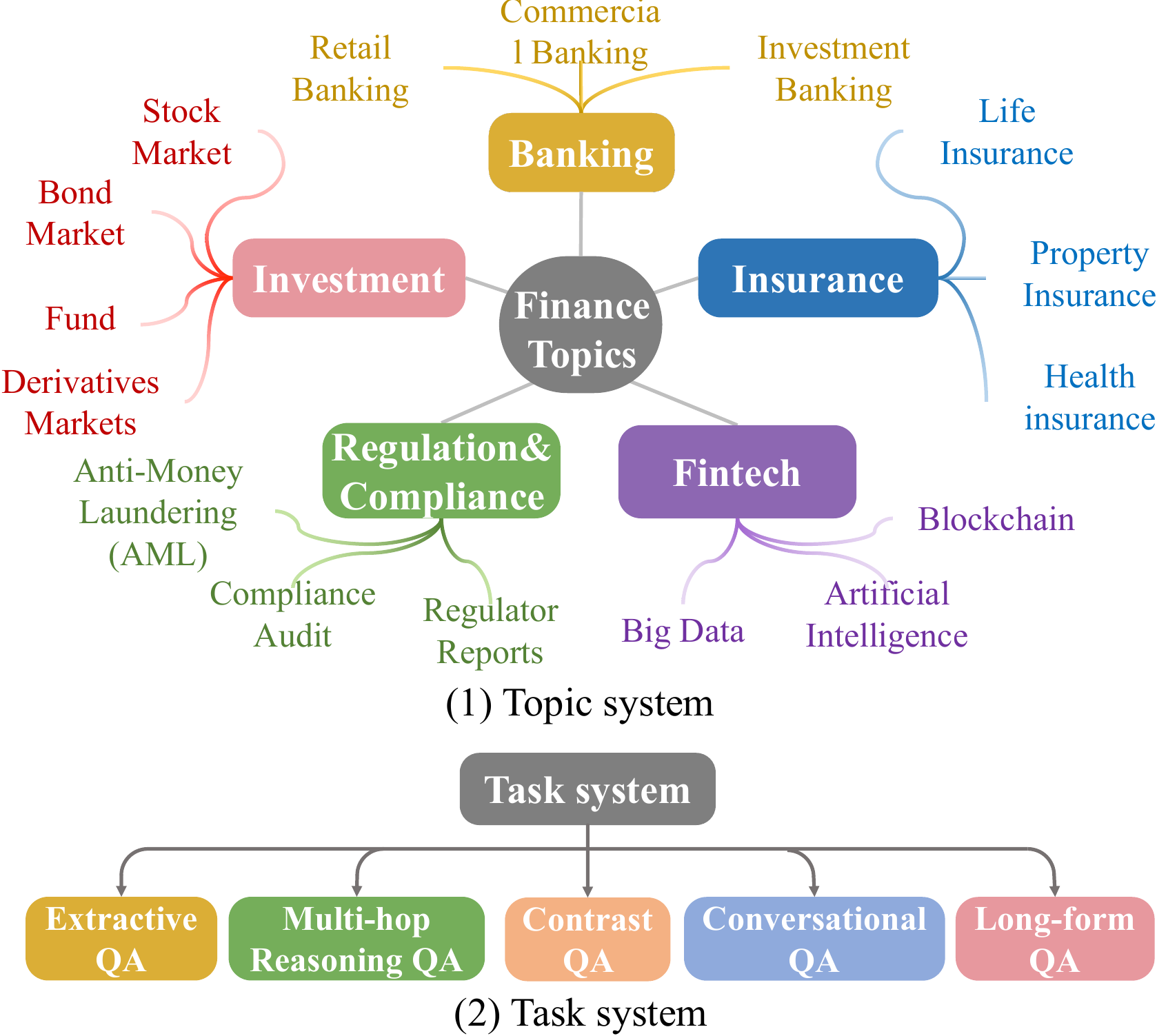}
    \caption{Topic \& task systems of our benchmark.}
    \label{fig:topic-tree}
\end{figure}

\subsection{Construction of Knowledge Corpus}\label{sec:knowledgecorpus}
To build a wide coverage and diverse financial document corpus, we collect our knowledge corpus from various data sources, including two open-source financial challenges, 
\href{https://www.modelscope.cn/datasets/BJQW14B/bs_challenge_financial_14b_dataset}{BS Challenge Financial}~(BSCF for short) and 
\href{https://tianchi.aliyun.com/competition/entrance/532164/introduction}{FinGLM};
finance-related web pages from \href{https://huggingface.co/datasets/wikimedia/wikipedia}{wikipedia-zh}; 
open-source financial pretraining dataset; \href{https://huggingface.co/datasets/BAAI/IndustryCorpus\_finance}{BAAI IndustryCorpus Finance (zh)}~(BAAI-Fin for short); 
and crawled financial web pages from the official agency websites. 
Considering that these external documents have various formats, such as PDF and SQLite, we use LlamaIndex\footnote{\href{https://www.llamaindex.ai/}{https://www.llamaindex.ai/}}, which is compatible with various data formats, to build our retrieval corpus. Specifically, we first transfer SQLite data to the JSON format, then utilize the LlamaIndex toolkit to split all documents into passages with the length set as 2048 and the overlap as 256. The statistical information of our data resources is shown in Table~\ref{tab:statistic_data_source}, where ``document'' denotes the LlamaIndex node.
\begin{figure*}[t]
    \centering
    \includegraphics[width=1\linewidth]{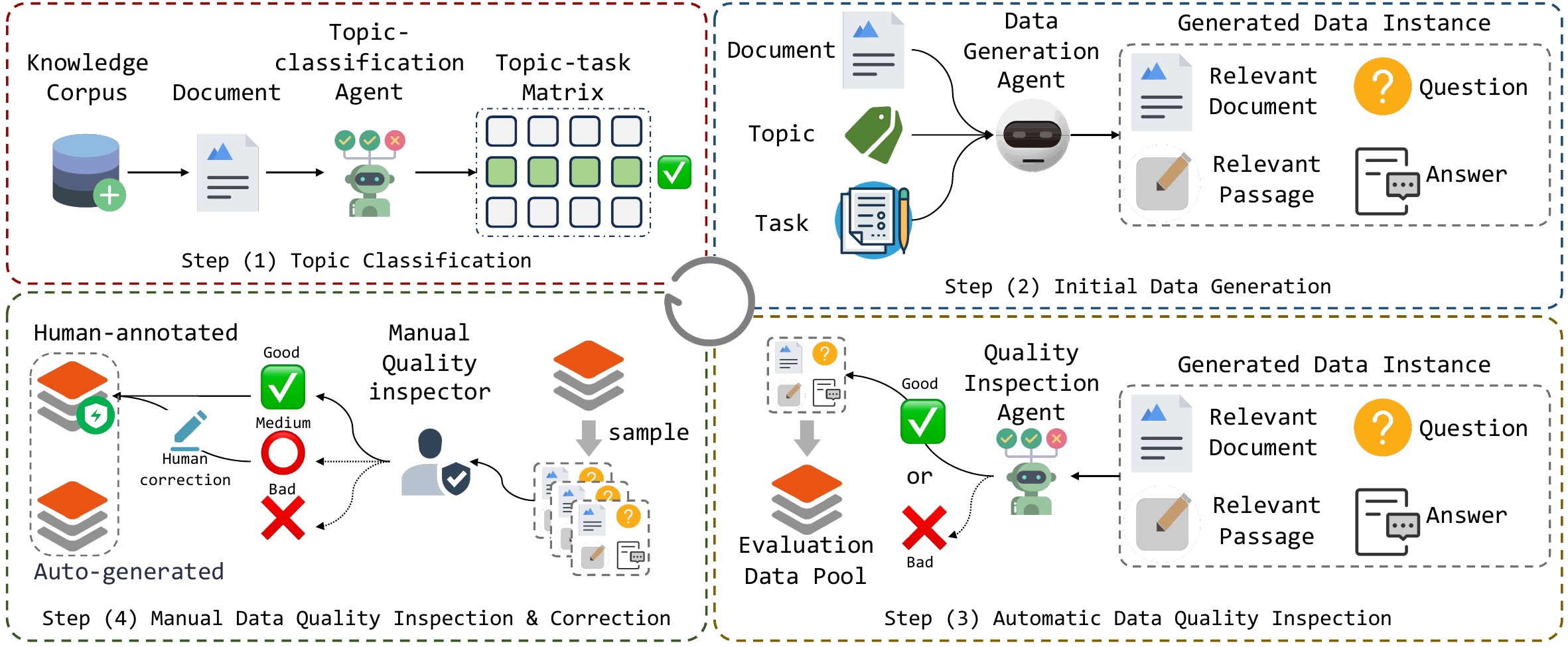}
    \caption{The visualization of \ours{}'s generation pipeline of evaluation data.}
    \label{fig:framework}
\end{figure*}

\subsection{Generation of Evaluation Instances}\label{sec:datagen}
Given the knowledge corpus with abundant domain-specific information, we set up our automatic data generation pipeline by a multi-agent method, supported by GPT-4. The processing steps of this pipeline are visualized in Figure~\ref{fig:framework}. 

\paragraph{RAG Scenario Recognition}
To construct matrix-based RAG evaluation scenarios that reflect real-world RAG applications, we classify our evaluation RAG scenarios from two orthogonal perspectives: domain topics and RAG tasks. 

From the topic perspective, we categorize RAG scenarios by domain topics related to user queries, such as the stock market and investment banks. Our topic system is initially generated from GPT-4, and we subsequently prune it according to the topic frequency. 
From the task perspective, we adopt five common and important RAG tasks, following existing studies~\cite{DomainRAG}: Extractive QA: Answers to queries can be extracted from the relevant documents without additional reasoning. Multi-hop reasoning QA: It requires multi-hop reasoning as answers are not explicitly stated in external documents. Contrast QA: It involves comparing two objects, requiring multi-aspect external knowledge to produce the final answer. Long-form QA: The queries demand detailed and comprehensive answers, which are usually long-form. Conversational QA: Answering the current question needs to consider the context of conversation histories.

The Cartesian product of these two perspectives forms an RAG scenario matrix, where each element represents a specific topic-task scenario. The topic and task systems used in our benchmark are presented in Figure~\ref{fig:topic-tree}. 
With the pre-defined topic-task matrix (\TM{}), we develop a \textit{topic classification agent} powered by GPT-4. This agent receives a sampled document from the knowledge corpus and then classifies the most relevant domain topic. This process locates a specific ``row'' in \TM{}. 
Subsequently, given the sampled document and the assigned topic, we will traverse all pre-defined RAG tasks to generate associated data instances for each RAG scenario within \TM{} elements. The generation approaches are demonstrated below. 

\paragraph{Data Generation}
Leveraging LLMs for automatic data generation and annotation has proven to be effective and reliable, significantly reducing the cost of human annotation~\cite{LLM-Anno-Survey}. In this context, we build a \textit{data generation agent} powered by GPT-4 to automatically generate data instances for our various RAG scenarios. 
Specifically, given a document, its domain topic, and a task description, we input these into the data generation agent to synthesize a question-answer pair. This pair is required to align with the task requirements and remain relevant to the topic. The input document is viewed as the relevant document of this QA pair. Additionally, to address the challenge of lengthy documents with extraneous information, we instruct the agent to identify the most relevant passage within the document, hence precisely locating the valuable content. As a result, each data instance comprises a user question, its answer, the relevant document, and a relevant passage.

\paragraph{Data Quality Inspection}
To ensure the quality of generated data instances, we develop a \textit{quality inspection agent} to filter out low-quality examples. The rationale behind this approach is that judging the instance quality is generally easier than generating high-quality data from scratch. Therefore, the inspection process could potentially improve the quality of the filtered dataset. This agent treats the generated data instance as input and predicts whether it contains meaningful information and meets the description of the target task. We only retain those instances that the quality inspection agent identifies as high-quality ones.

\paragraph{Manual Quality Inspection and Correction}
Besides agent-based quality inspection, we employ annotators to perform data quality inspection and correction, leading to a high-quality evaluation dataset and enhanced reliability of our benchmark.

We first sample a subset from generated instances for each \TM{} element. Annotators are then requested to check the following aspects of the data:
Does the generated question meet the \textit{task requirements}? Is the question \textit{related to the given topic}? Is the question \textit{semantically complete}? Is the \textit{answer correct and complete}?. Are the \textit{extracted passages accurate and complete}? 
The annotation follows a five-point scale from 1 to 5, where 1 and 2 indicate low data quality, suggesting that the instance should be discarded; 3 signifies the data contains some human-fixable defects; and 4 or 5 denotes good to excellent data quality. The number of labeled data instances is 910. 

\begin{figure}
    \centering
    \includegraphics[width=1\linewidth]{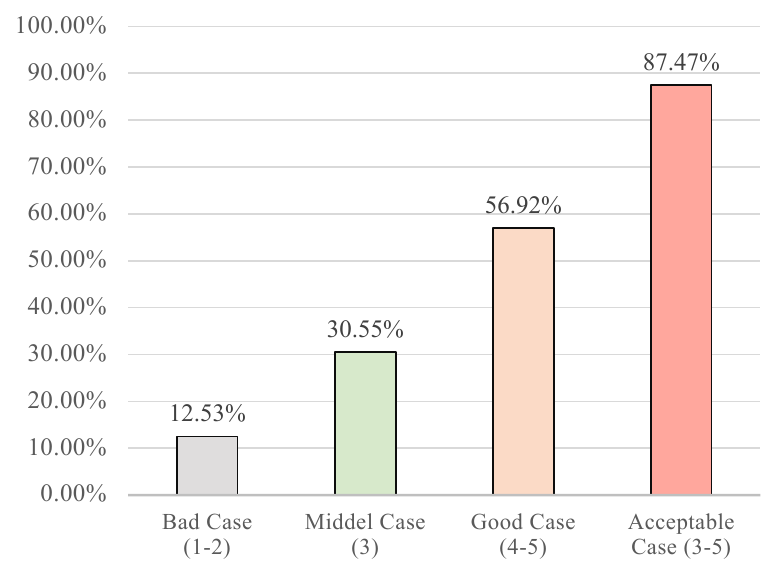}
    \caption{Statistical information of manual inspection.}
    \label{fig:human_static}
\end{figure}

We present the statistical results of the inspection in Figure~\ref{fig:human_static}. The findings reveal that the acceptance rate of our auto-generated cases is 87.47\%, potentially confirming the effectiveness and usability of our multi-agent-based data generation pipeline. Annotators are also tasked with correcting instances labeled as 3 to create high-quality human-annotated data.
Through these inspection and correction steps, we establish a reliable human-annotated dataset, significantly enhancing the robustness of our benchmark.  
Finally, we create two datasets: one auto-generated and the other human-annotated. We further split the auto-generated ones into train and test datasets to facilitate related investigations based on our benchmark. 

The data amounts of these datasets are shown in Appendix~\ref{app:data-statistic} and the instructions we used for GPT-4 and annotators are shown in Appendix~\ref{app:instructions}.

\subsection{Evaluation of RAG Models}\label{sec:evaluation}
To comprehensively and accurately assess RAG baselines, we integrate two types of metrics: rule-based metrics and model-based metrics.

\paragraph{Rule-based Metrics}
Given the widespread usage and stability of rule-based metrics, we use Rouge-L to provide foundational evaluations for RAG systems.\footnote{\href{https://pypi.org/project/rouge-chinese/}{https://pypi.org/project/rouge-chinese/}} We also incorporate ranking metrics, MAP and MRR, to assess the performance of retrievers within RAG systems. This combination facilitates a holistic evaluation of the entire RAG pipeline.

\paragraph{Model-based Metrics}
Given the flexibility and diversity of AI chatbot responses, rule-based metrics often struggle to provide semantic evaluations. To solve it, we devise five high-level metrics implemented based on fine-tuned LLMs:

% \noindent$\bullet$ Accuracy (ACC).
% \noindent
\textit{Accuracy (ACC).}
LLMs often generate responses that are correct in content but poorly matched in wording. Therefore, we propose a model-based accuracy metric to measure semantic alignment between LLM responses and golden answers. It is a three-scale metric, where 1 indicates poor quality, 2 means average quality, and 3 is good quality.

\begin{table}[!t]
    \centering
    \resizebox{0.99\linewidth}{!}{\begin{tabular}{llcc}
        \toprule
        Setting & Base Model & $\kappa$ & Accuracy \\
        \midrule
        Prompting & Llama3.1-8B-Inst & 39.70 & 55.60  \\
        Prompting & Llama3.1-70B-Inst & 54.14 & 66.40  \\
        Prompting & Qwen2.5-7B-Inst  & 48.05 & 62.00  \\
        Prompting & Qwen2.5-32B-Inst & \underline{61.44} & \underline{71.60} \\
        Prompting & Qwen2.5-72B-Inst & 55.38 & 67.20  \\
        \midrule
        Lora-FT & Llama3.1-8B-Inst  & 48.63  & 62.80  \\
        Lora-FT & Qwen2.5-7B-Inst	 & \textbf{64.86} & \textbf{74.40}  \\
        \bottomrule
    \end{tabular}}
    \caption{Experimental results of model-based evaluator.}
    \label{tab:evaluator_res}
\end{table}

% \noindent$\bullet$ Completeness (COM).
% \noindent
\textit{Completeness (COM).}
Long-form QA usually requires LLM to provide comprehensive answers that address various aspects of the question~\cite{RichRAG}. To assess completeness, we introduce a four-point metric: 1 indicates the response hits no relevant aspects to the question; 2 signifies the response partially satisfies relevant aspects; 3 means the response covers all aspects comprehensively; and -1 indicates that completeness measurement is not applicable for the input QA scenario.

% \noindent$\bullet$ Hallucination (HAL).
% \noindent
\textit{Hallucination (HAL).}
It assesses hallucinations in generated responses: HAL is 0 if the response is correct, or incorrect but derived from retrieved documents; HAL is 1 if the response is incorrect and unrelated to the retrieved content; and HAL is -1 if hallucination evaluation is unnecessary. 

% \noindent$\bullet$  Utilization (UTL).
% \noindent
\textit{Utilization (UTL).}
This metric assesses whether LLMs effectively utilize retrieved documents and whether the answer could be traced from retrieved documents. Similarly to ACC, it is also three-scale.

% \noindent$\bullet$  Numerical accuracy (NAC).
% \noindent
\textit{Numerical accuracy (NAC).}
This metric addresses scenarios involving financial computations, where answers are typically numerical. It is a three-scale metric: 1 indicates correct, 0 means wrong, and -1 means the answer is non-numerical. 

Finally, all metrics are normalized into [0,1], and samples evaluated as -1 will not be considered for the specific metrics.

\paragraph{SFT of LLM evaluator}
To ensure the reliability of our LLM evaluator, we conduct human annotation on a subset of generated responses for the five metrics, creating a labeled dataset for training stable evaluators. Specifically, we randomly sample 127 cases and produce 635 examples by aggregating all five metrics. We divide it into training, validation, and test sets in a ratio of 5:1:4. 

Leveraging the robust capabilities of LLMs, we observe distinct improvements in evaluation performance, even with limited training data. We experiment with prompting and Lora~\cite{lora} fine-tuning on Qwen2.5 and Llama3.1 across various model sizes. Results are presented in Table~\ref{tab:evaluator_res} with accuracy and $\kappa$ value as evaluation metrics, measuring the agreement with ground truths. 
Finally, we build our evaluator by the fine-tuned Qwen-2.5-7B-Instruct with the best performance.

\section{Experiment}
We conduct our experiments on various open-resource retrievers and LLMs. Specifically, for \textbf{retrievers}, we select GTE-Qwen2-1.5B~\cite{gte}, BGE-large-zh~\cite{bge_embedding}, BGE-M3~\cite{bge_embedding}, and Jina-zh~\cite{jina}. For \textbf{LLMs}, we choose Qwen2.5-72B-Instruct~\cite{qwen2.5}, Deepseek-v2-chat~\cite{deepseekv2}, Yi15-34b~\cite{yi}, and Llama3.1-70B-Instruct~\cite{llama31}. In our experiments, we set the retrieved document number as 5 to ensure a fair comparison. 
\begin{table*}[!ht]
\small
    \centering
    \resizebox{0.99\linewidth}{!}{\begin{tabular}{lccccccccc}
    \toprule
    Models & MAP $\uparrow$ & MRR $\uparrow$ & Rouge-L $\uparrow$ & F1 $\uparrow$ & ACC $\uparrow$ & HAL $\downarrow$ & COM  $\uparrow$ & UTL  $\uparrow$  & NAC  $\uparrow$  \\
    \midrule
    \multicolumn{10}{c}{Auto-generated evaluation set}\\
    \midrule
    % \multicolumn{10}{c}{Auto-generated evaluation set} 
    Jina-zh & 0.3395 & 0.3469 & 0.1662 & 0.2553 & 0.3908 & 0.0794 & 0.5981 & 0.5078 & 0.2837 \\
    BGE-large-zh & 0.3777 & 0.3865 & 0.1693 & 0.2541 & 0.4080 & \underline{0.0597} & 0.6048 & 0.5194 & \underline{0.3124} \\
    BGE-M3 & \underline{0.3961} & \underline{0.4057} & \underline{0.1746} & \textbf{0.2593} & \underline{0.4091} & 0.0634 & \underline{0.6092} & \underline{0.5203} & 0.3060 \\
    GTE-Qwen2-1.5B & \textbf{0.4370} & \textbf{0.4491} & \textbf{0.1778} & \underline{0.2563} & \textbf{0.4326} & \textbf{0.0467} & \textbf{0.6256} & \textbf{0.5613} & \textbf{0.3293} \\
    \midrule
    \multicolumn{10}{c}{Human-annotated evaluation set}\\
    \midrule
    Jina-zh & 0.3458 & 0.3533 & 0.2341 & 0.3821 & 0.4089 & 0.0886 & 0.5930 & 0.5163 & 0.3073 \\
    BGE-large-zh & \underline{0.4153} & \underline{0.4252} & 0.2435 & 0.3870 & 0.4325 & 0.0718 & \textbf{0.6224} & 0.5367 & \underline{0.3545} \\
    BGE-M3 & 0.4152 & 0.4236 & \underline{0.2517} & \underline{0.3913} & \underline{0.4450} & \underline{0.0709} & \underline{0.6208} & \underline{0.5410} & 0.3472 \\
    GTE-Qwen2-1.5B & \textbf{0.4443} & \textbf{0.4574} & \textbf{0.2528} & \textbf{0.3919} & \textbf{0.4476} & \textbf{0.0618} & 0.6190 & \textbf{0.5576} & \textbf{0.3595} \\
    \bottomrule
    \end{tabular}}
    \caption{The overall results of retrieval models with the generator being set as Qwen2.5-72B.}
    \label{tab:ret_res}
\end{table*}

\begin{table*}[!h]
    \centering
    \resizebox{0.99\linewidth}{!}{\begin{tabular}{llccccccc}
    \toprule
    Retriever & Generator & Rouge-L $\uparrow$ & F1 $\uparrow$ & ACC $\uparrow$ & HAL $\downarrow$ & COM  $\uparrow$ & UTL  $\uparrow$  & NAC  $\uparrow$  \\
    \midrule
    \multicolumn{9}{c}{Auto-generated evaluation set}  \\
    \midrule
    Close-Book & Yi15-34B & 0.0326 & 0.0673 & 0.1573 & - & 0.5063 & - & 0.0693 \\
    Close-Book & Deepseek-v2-chat & 0.1861 & 0.3709 & 0.3587 & - & 0.5755 & - & 0.1121 \\
    Close-Book & Qwen2.5-72B & 0.1607 & 0.3222 & 0.3788 & - & 0.6017 & - & 0.1256 \\
    Close-Book & Llama3.1-70B-Instruct & 0.1993 & 0.3989 & 0.3238 & - & 0.5284 & - & 0.0677 \\

    GTE-Qwen2-1.5B & Yi15-34B & 0.0593 & 0.0958 & 0.3402 & \underline{0.0597} & 0.5778 & 0.4229 & 0.1682 \\
    GTE-Qwen2-1.5B & Deepseek-v2-chat & \underline{0.2279} & \underline{0.3300} & 0.4099 & 0.0634 & \underline{0.6072} & \underline{0.5197} & \underline{0.3175} \\
    GTE-Qwen2-1.5B & Qwen2.5-72B & 0.1778 & 0.2563 & \underline{0.4326} & \textbf{0.0467} & \textbf{0.6256} & \textbf{0.5613} & \textbf{0.3293} \\
    GTE-Qwen2-1.5B & Llama3.1-70B-Instruct & \textbf{0.3235} & \textbf{0.4810} & \textbf{0.4398} & 0.0792 & 0.5926 & 0.4754 & 0.3088 \\
    \midrule
    \multicolumn{9}{c}{Human-annotated evaluation set}  \\
    \midrule
    Close-Book & Yi15-34B & 0.0497 & 0.1161 & 0.1461 & - & 0.4987 & - & 0.0749 \\
    Close-Book & Deepseek-v2-chat & 0.2250 & 0.4353 & 0.3306 & - & 0.5541 & - & 0.1153 \\
    Close-Book & Qwen2.5-72B & 0.2082 & 0.4191 & 0.3405 & - & 0.5754 & - & 0.1241 \\
    Close-Book & Llama3.1-70B-Instruct & 0.2195 & 0.4183 & 0.2859 & - & 0.5133 & - & 0.0659 \\

    GTE-Qwen2-1.5B & Yi15-34B & 0.0887 & 0.1583 & 0.3366 & \underline{0.0648} & 0.5821 & 0.4234 & 0.1856 \\
    GTE-Qwen2-1.5B & Deepseek-v2-chat & \underline{0.2916} & \underline{0.4353} & 0.4234 & 0.0750 & \underline{0.6006} & \underline{0.5160} & 0.3213 \\
    GTE-Qwen2-1.5B & Qwen2.5-72B & 0.2528 & 0.3919 & \textbf{0.4476} & \textbf{0.0618} & \textbf{0.6190} & \textbf{0.5576} & \textbf{0.3595} \\
    GTE-Qwen2-1.5B & Llama3.1-70B-Instruct & \textbf{0.3390} & \textbf{0.5042} & \underline{0.4433} & 0.1131 & 0.5745 & 0.4764 & \underline{0.3268} \\
    \bottomrule
    \end{tabular}}
    \caption{The overall evaluation results on final responses of RAG models.}
    \label{tab:gen_res}
\end{table*}

\begin{figure*}[!h]
\centering
    \includegraphics[width=1\linewidth]{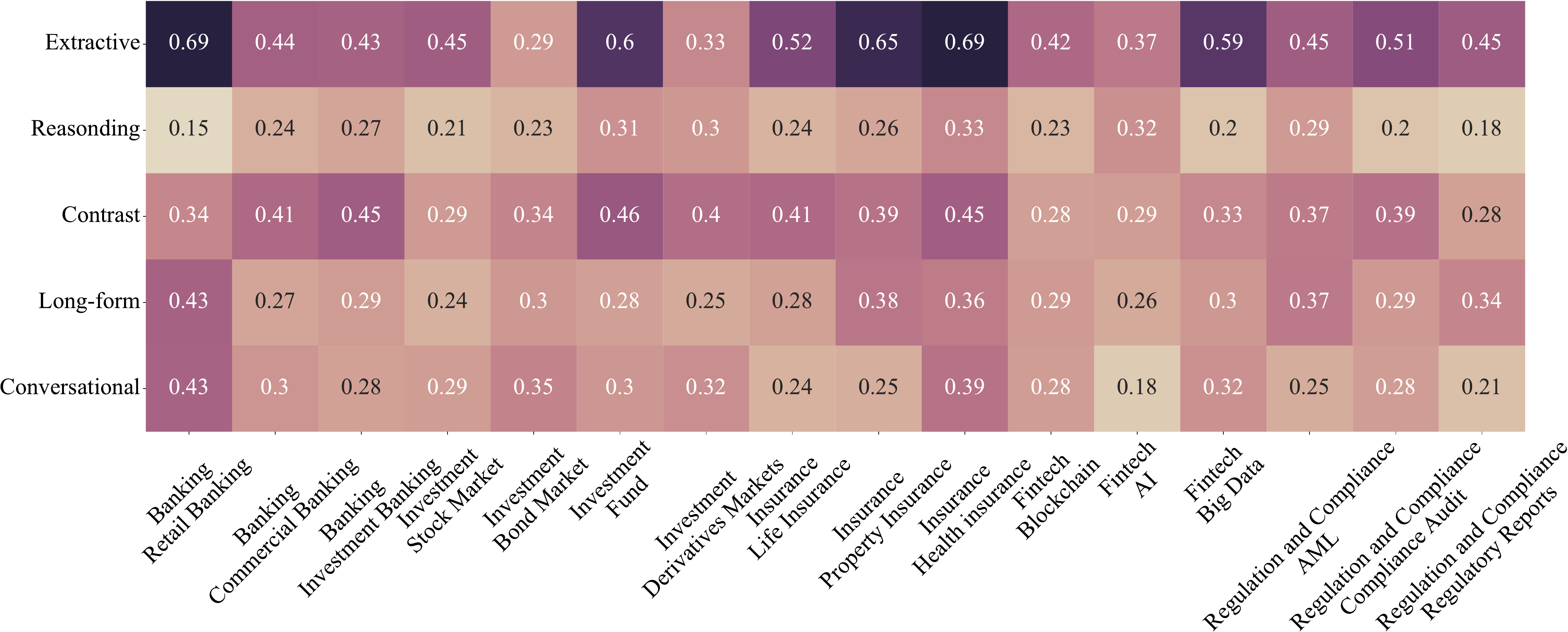}
    \caption{Rouge-L of matrix-based results of GTE-Qwen2-1.5B+Llama3.1-70B-Instruct on human-annotated subsets.}
    \label{fig:gte-qwen2-1.5b_TOP5-llama3-70b-instruct}
\end{figure*}

\subsection{Comparison Experiments of Retrievers}
Our experiments aim to assess the entire pipeline of RAG systems, including both retrievers and generators (LLMs). First, we present the experimental results on retrievers using our two evaluation datasets, the auto-generated set and the human-annotated set, with the generator set as Qwen2.5-72B. 

The main results are displayed in Table~\ref{tab:ret_res}. According to the results shown, GTE-Qwen2-1.5B demonstrates the best retrieval performance across most retrieval and generation metrics. We attribute this superiority to two factors: (1) Model parameters:  GTE-Qwen2-1.5B encompasses the most model parameters among all baselines, significantly enhancing its performance upper bound. (2) Fine-tuning from LLM: It is continuously fine-tuned from the LLM, Qwen2-1.5B, which is pre-trained using a large-scale corpus. This strategy equips it with extensive world knowledge, providing better prior knowledge compared to retrievers that are pre-trained from scratch.

\subsection{Comparison Experiments of Generators}
Next, we evaluate the abilities of generators to solve expert finance-related problems. Given the superiority of GTE-Qwen2-1.5B in the retrieval task, we choose it as our retriever and compare the response quality of selected popular LLMs. The main results are presented in Table~\ref{tab:gen_res}. In this context, the setting ``Close-Book'' indicates that responses are generated solely by LLMs without incorporating retrieved external knowledge. Since HAL and UTL metrics are required to be evaluated based on the retrieved results, there are no corresponding results in the close-book settings. 

Based on the results, we conclude the following findings: (1) RAG systems generally outperform close-book LLMs on our evaluation datasets. We notice that LLMs typically yield better results when equipped with retrievers compared to close-book settings. It proves that in domain-specific scenarios, it is essential for LLM to retrieve external expert knowledge, thereby enhancing the reliability of generated responses. (2) There remains significant potential for existing retrievers and LLMs to enhance RAG abilities in financial domains. Even with the RAG systems, performance is still lacking across all retriever and LLM configurations. This indicates the difficulty of our evaluation datasets, which involve expert and reasoning financial tasks. Additionally, it confirms that our benchmark introduces new challenges for existing RAG systems, potentially driving further investigation into RAG models in domain-specific scenarios.

\subsection{Experiments on Topic-specific Subsets}
\begin{figure}
    \centering
    \includegraphics[width=1\linewidth]{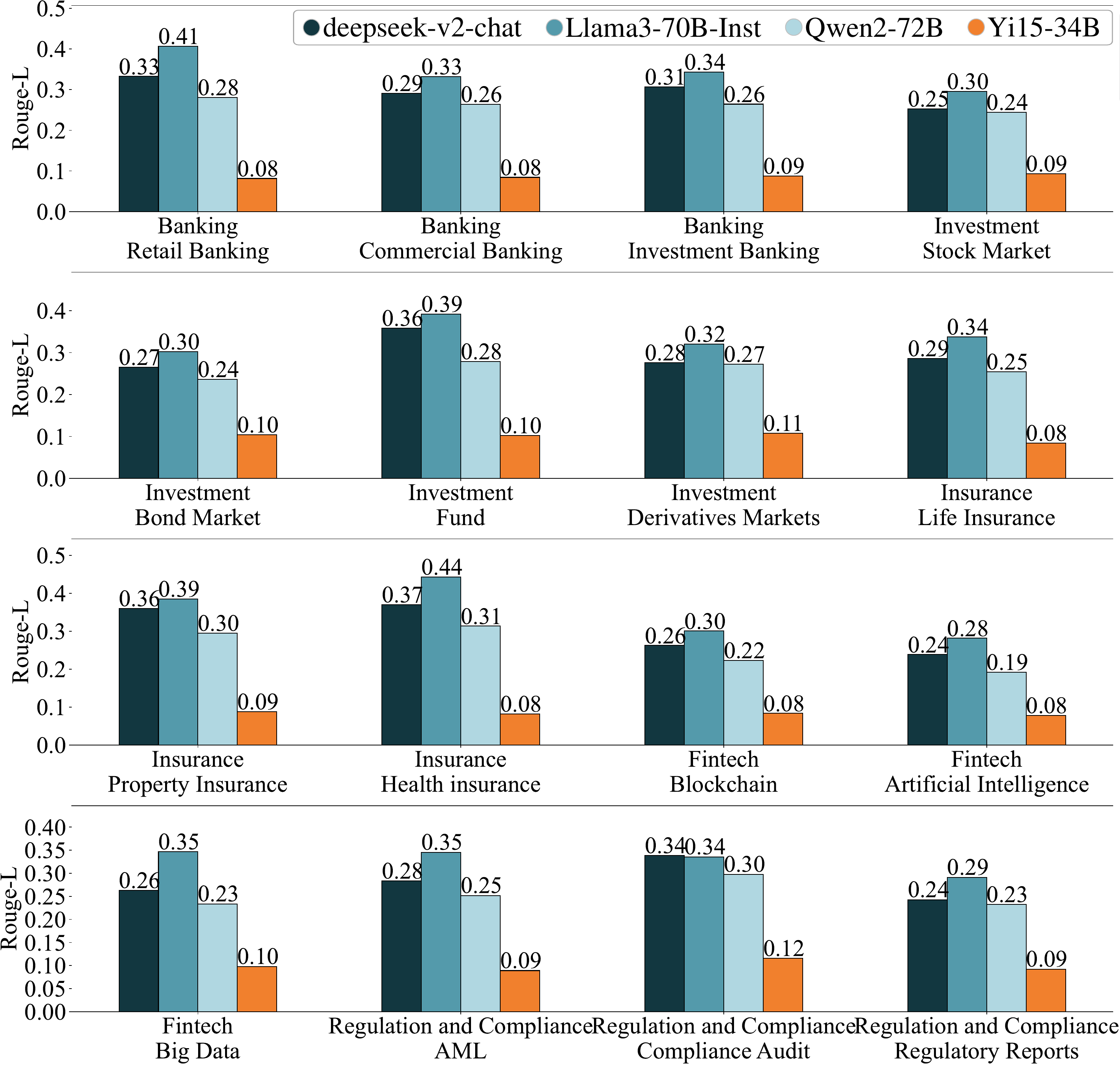}
    \caption{Rouge-L scores of generators on topic-specific human-annotated subsets.}
    \label{fig:topic-set-rougel}
\end{figure}

As previously mentioned, we build a topic tree to create several subsets, thereby evaluating RAG systems across different scenarios with diverse query topics. We further demonstrate the performance of RAG models on these topic-specific subsets to clearly demonstrate their abilities to handle various topic scenarios. The results are illustrated in Figure~\ref{fig:topic-set-rougel}. Due to limited space, we present the topic-specific results on auto-generated sets in Appendix~\ref{app:exp-figure}, \ie, Figure~\ref{fig:topic-autoset-rougel}.

We notice that the same RAG model exhibits varying performance across different topic scenarios, indicating an imbalance in their capabilities to solve different query scenarios. This inconsistency may arise from the different popularity of topics within the pre-trained corpus of LLMs, leading to imbalanced RAG abilities. Consequently, how to balance the capabilities of RAG models across diverse topics with varied popularities may also be an important investigation direction.

\subsection{Experiments on Task-specific Subsets}
Utilizing our \TM{}-based evaluation subsets, we further compare RAG models across different task evaluation sets, assessing their abilities on diverse query tasks. The experimental results are illustrated in Figure~\ref{fig:task-set-rougel}. Due to limited space, we present the task-specific results on auto-generated sets in Appendix~\ref{app:exp-figure}, \ie, Figure~\ref{fig:task-autoset-rougel}.

It is evident that the performance of the RAG system varies significantly across different query tasks. This phenomenon may stem from the differing difficulty levels of these tasks. For example, most RAG models perform poorly on multi-hop reasoning and conversational QA tasks. It is because these tasks require robust reasoning and context-understanding abilities, making it more challenging for RAG models to generate accurate responses. Thus, investigating ways to enhance RAG systems in these challenging but practical tasks also represents a promising and important research direction.

\subsection{Matrix-based Visualization of Results}
As we mentioned earlier, our matrix-based evaluation scenarios offer a comprehensive ability profile for the evaluated RAG model, distinctly revealing their performance on specific topic-task scenarios. Accordingly, we present a representative matrix-based visualization of GTE-Qwen2-1.5B+Llama3.1-70B-Instruct on human-annotated subsets, which is shown in Figure~\ref{fig:gte-qwen2-1.5b_TOP5-llama3-70b-instruct}. Due to limited spaces, we illustrate the results of other models on auto-generated and human-annotated subsets in Appendix~\ref{app:exp-figure}, \ie, Figures~\ref{fig:gte-qwen2-1.5b_TOP5-llama3-70b-instruct-auto},~\ref{fig:gte-qwen2-1.5b_TOP5-qwen-human},~\ref{fig:gte-qwen2-1.5b_TOP5-qwen-auto},~\ref{fig:gte-qwen2-1.5b_TOP5-deepseek-human}~\ref{fig:gte-qwen2-1.5b_TOP5-deepseek-auto},~\ref{fig:gte-qwen2-1.5b_TOP5-yi-human}, and~\ref{fig:gte-qwen2-1.5b_TOP5-yi-auto}.
\begin{figure}
    \centering
    \includegraphics[width=1\linewidth]{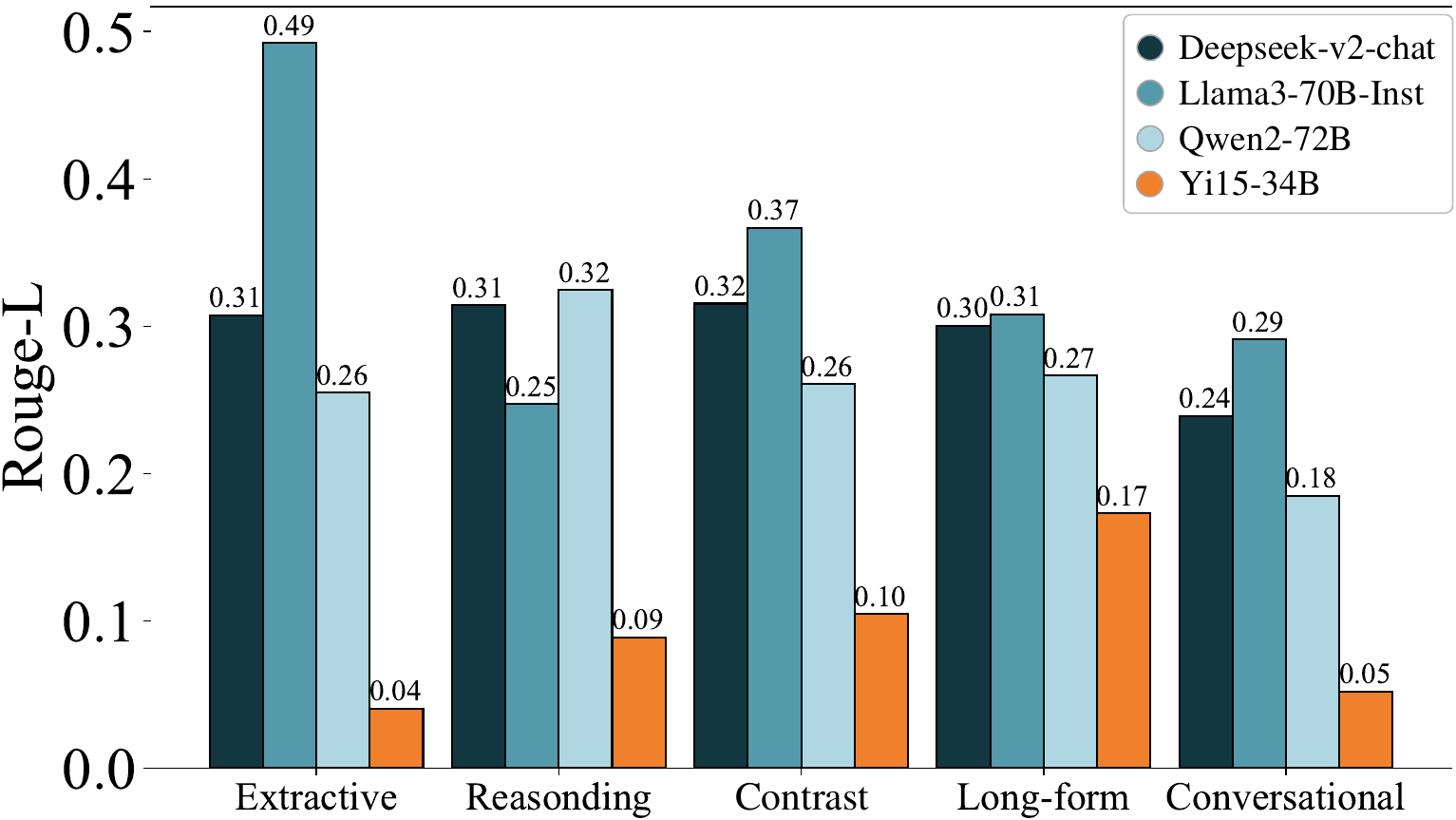}
    \caption{Rouge-L scores of generators on task-specific human-annotated subsets.}
    \label{fig:task-set-rougel}
\end{figure}
This method highlights the specific abilities of RAG models more clearly than simply averaging all results, allowing for more detailed and fine-grained analyses. For example, in Figure~\ref{fig:gte-qwen2-1.5b_TOP5-llama3-70b-instruct}, which presents the results of GTE-Qwen2-1.5B+Deepseek-v2, it is evident that this RAG model excels in the extractive QA task with the ``Fund''-related topic. However, there remains significant room for improvement in the conversational QA task with the ``AI''-related topic. Such visualization provides a novel approach to analyzing RAG performance across different scenarios, enabling targeted strategies to address the localized limitations of RAG models.

\section{Conclusion}
In this study, we propose an automatic and omnidirectional RAG benchmark in a vertical domain \ie, finance. We first identify diverse query scenarios via a matrix-based method, which considers two orthogonal perspectives, topics, and tasks. This approach allow us to assess RAG systems comprehensively and finely by simulating diverse practical RAG scenarios. We utilize the multi-agent technique to automatically construct our evaluation datasets. Through rigorous model-based and manual quality inspections, we derive three datasets: an auto-generated training set, an auto-generated test set, and a human-annotated test set. The high acceptance of auto-generated data confirms the reliability of our data generation methods. Our experimental results illustrate that there is still a significant improvement space for existing RAG models in vertical domains. In addition, RAG systems exhibit varying performance across diverse query scenarios, highlighting new challenges and investigation directions for RAG studies.

\section*{Limitations}
In this study, we develop an omnidirectional and automated RAG benchmark specifically tailored for the finance domain. 
Our benchmark is featured by its matrix-based RAG evaluation scenarios, multi-dimensional data generation approaches that combine automatic and manual methods, a multi-stage evaluation pipeline, and a multi-dimensional evaluation system. 
However, we acknowledge several limitations that warrant further investigation:

First, despite our efforts to collect a diverse data corpus, the distribution remains somewhat limited. This limitation arises primarily from challenges related to accessibility and the open licensing of data resources. As a result, there is a risk of introducing potential biases into our datasets, which could affect the generalizability of our benchmark findings. 
Second, we recognize that the costs associated with human annotation have led to a limited amount of collected human evaluation data for training our LLM evaluators, which may impact the performance of LLM evaluators. In future studies, we plan to gather a more extensive set of human evaluation data. This enhancement aims to boost the accuracy and reliability of our LLM evaluators, ultimately leading to a more effective benchmark.

% \newpage
\bibliography{custom}
% \newpage
\appendix
\begin{figure}[!h]
    \centering
    \includegraphics[width=1\linewidth]{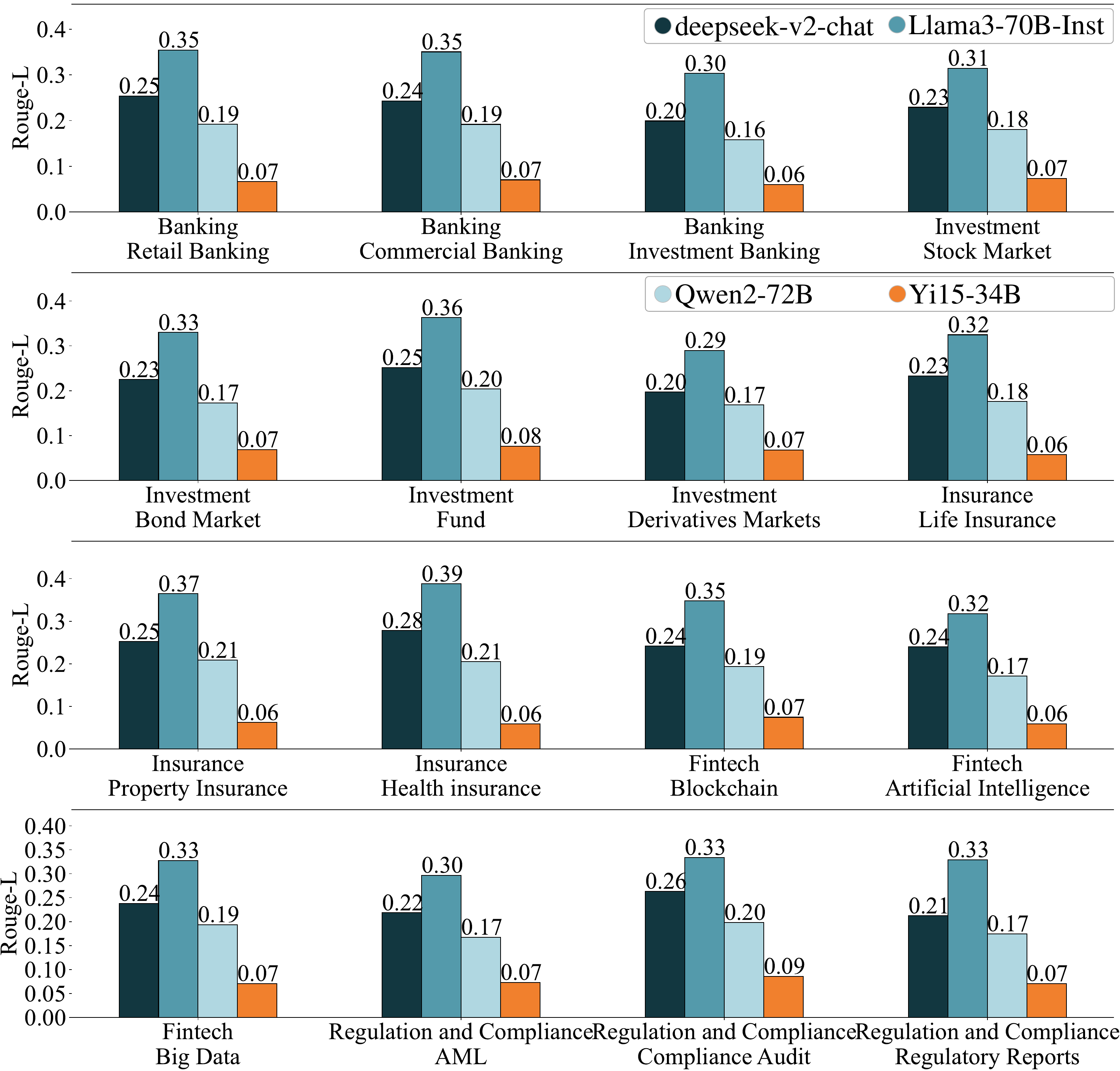}
    \caption{Rouge-L scores of generators on topic-specific auto-generated subsets.}
    \label{fig:topic-autoset-rougel}
\end{figure}
\begin{figure}[!h]
    \centering
    \includegraphics[width=1\linewidth]{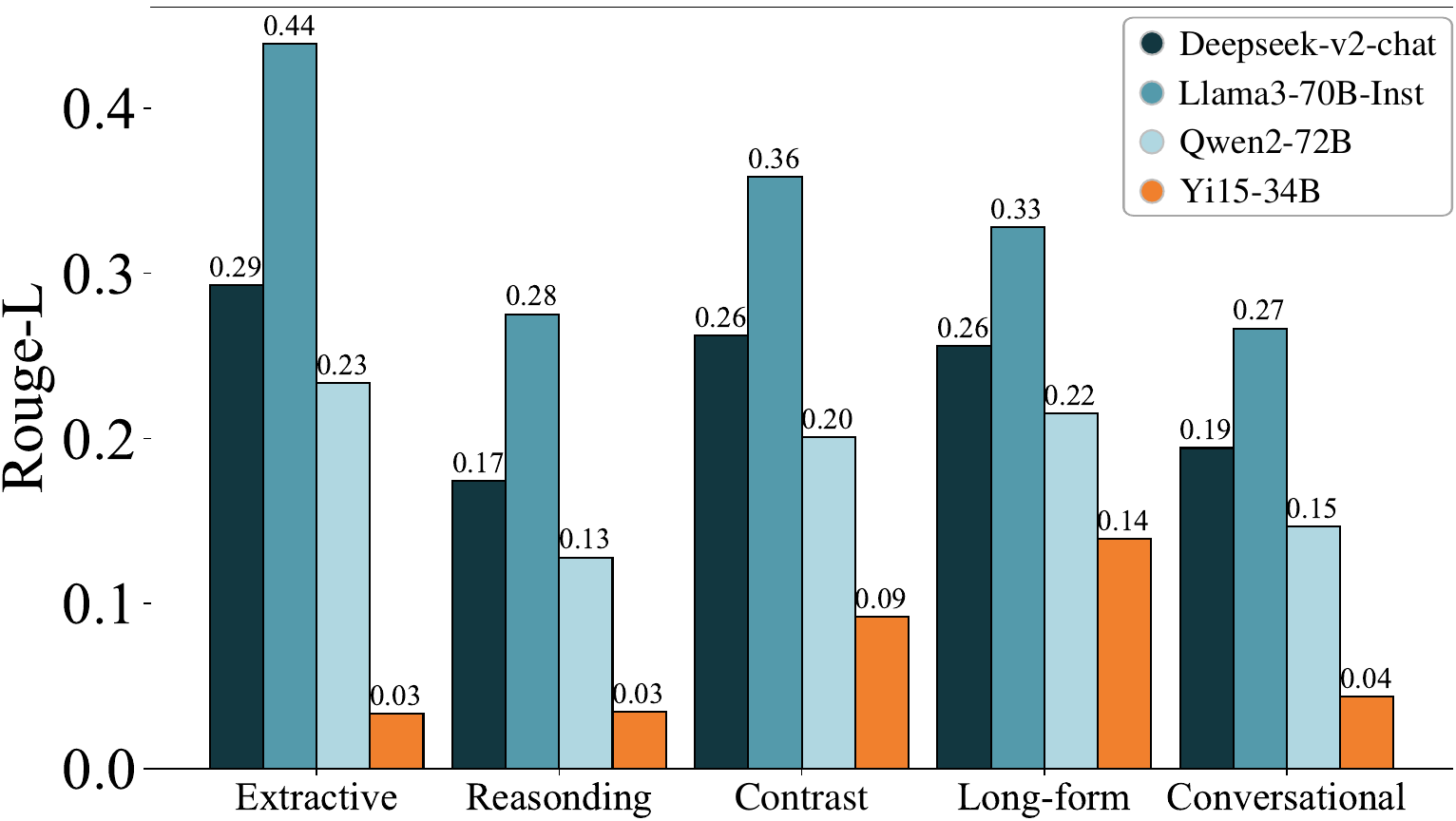}
    \caption{Rouge-L scores of generators on task-specific auto-generated subsets.}
    \label{fig:task-autoset-rougel}
\end{figure}

\section{Statistical Information of Our datasets}\label{app:data-statistic}
\begin{figure*}[!h]
    \includegraphics[width=1\linewidth]{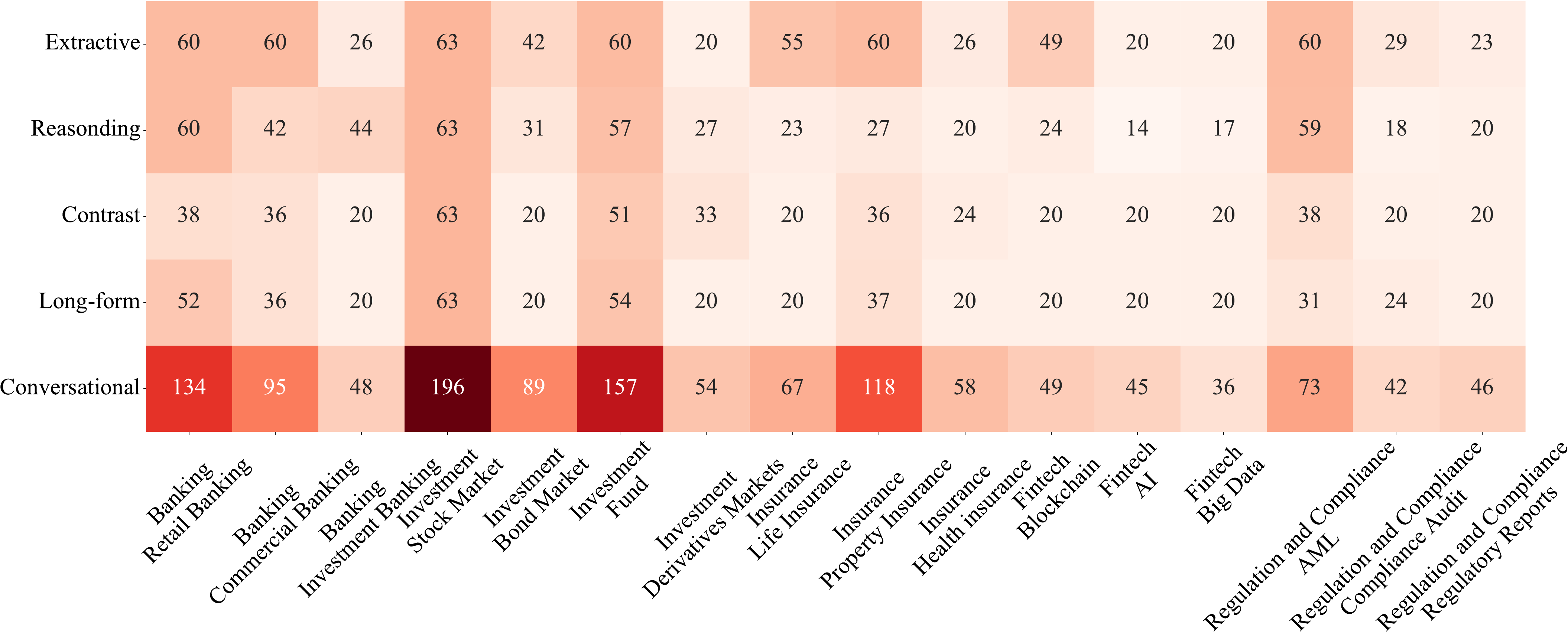}
    \caption{Data amount of the auto-generated training set.}
    \label{fig:train_count}
\end{figure*}

\begin{figure*}
    \includegraphics[width=1\linewidth]{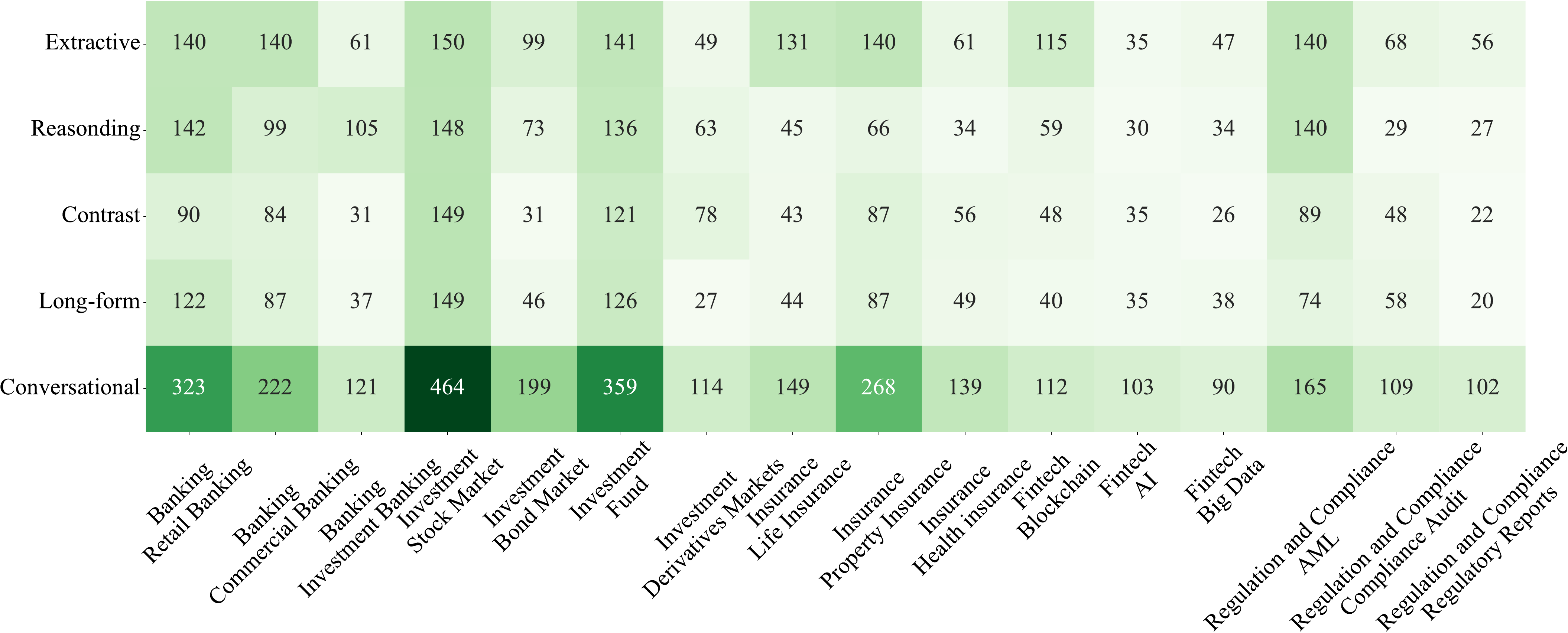}
    \caption{Data amount of the auto-generated test set.}
    \label{fig:test_count}
\end{figure*}

\begin{figure*}
    \includegraphics[width=1\linewidth]{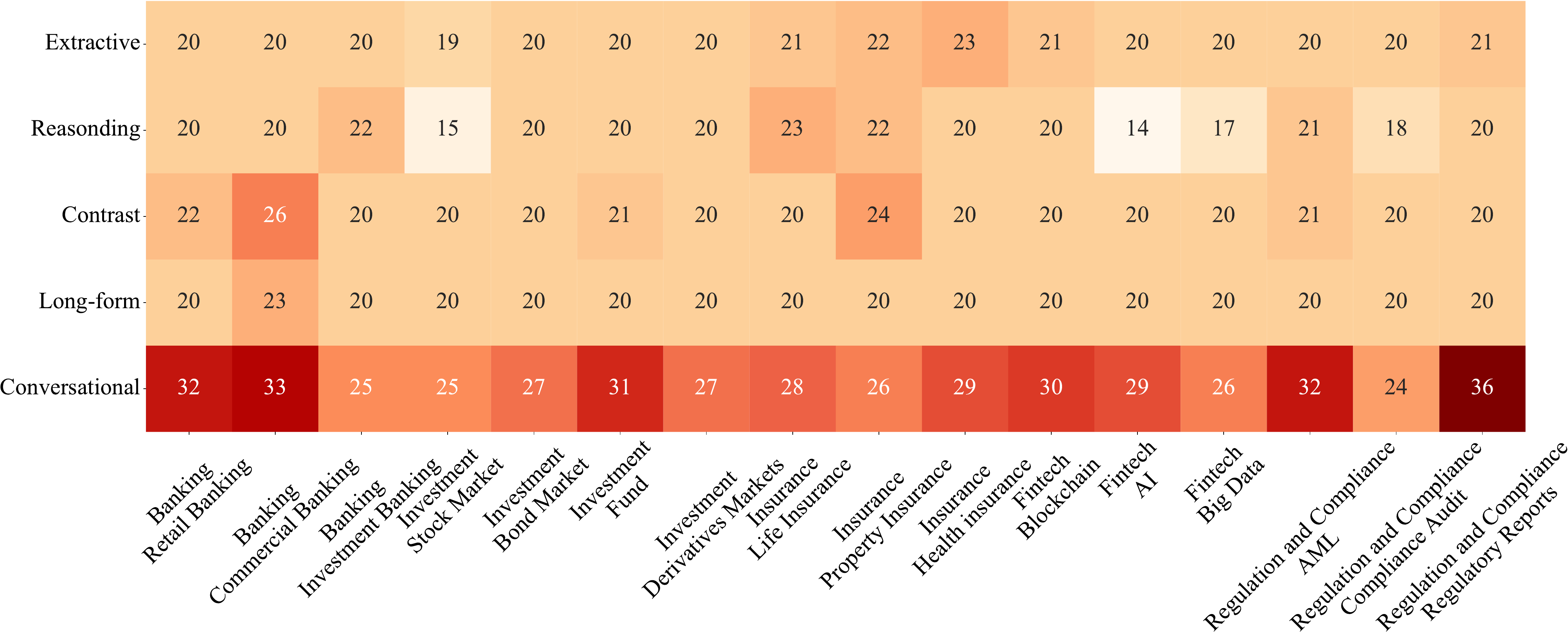}
    \caption{Data amount of the human-annotated test set.}
    \label{fig:human_count}
\end{figure*}

In this section, we provide the detailed statistical information of our three datasets, including auto-generated training set, auto-generated test set, and human-annotated test set, in Figure~\ref{fig:train_count},~\ref{fig:test_count}, and~\ref{fig:human_count}.


\section{Supplementary Visualization Results}\label{app:exp-figure}
\begin{figure*}[!h]
\centering
    \includegraphics[width=1\linewidth]{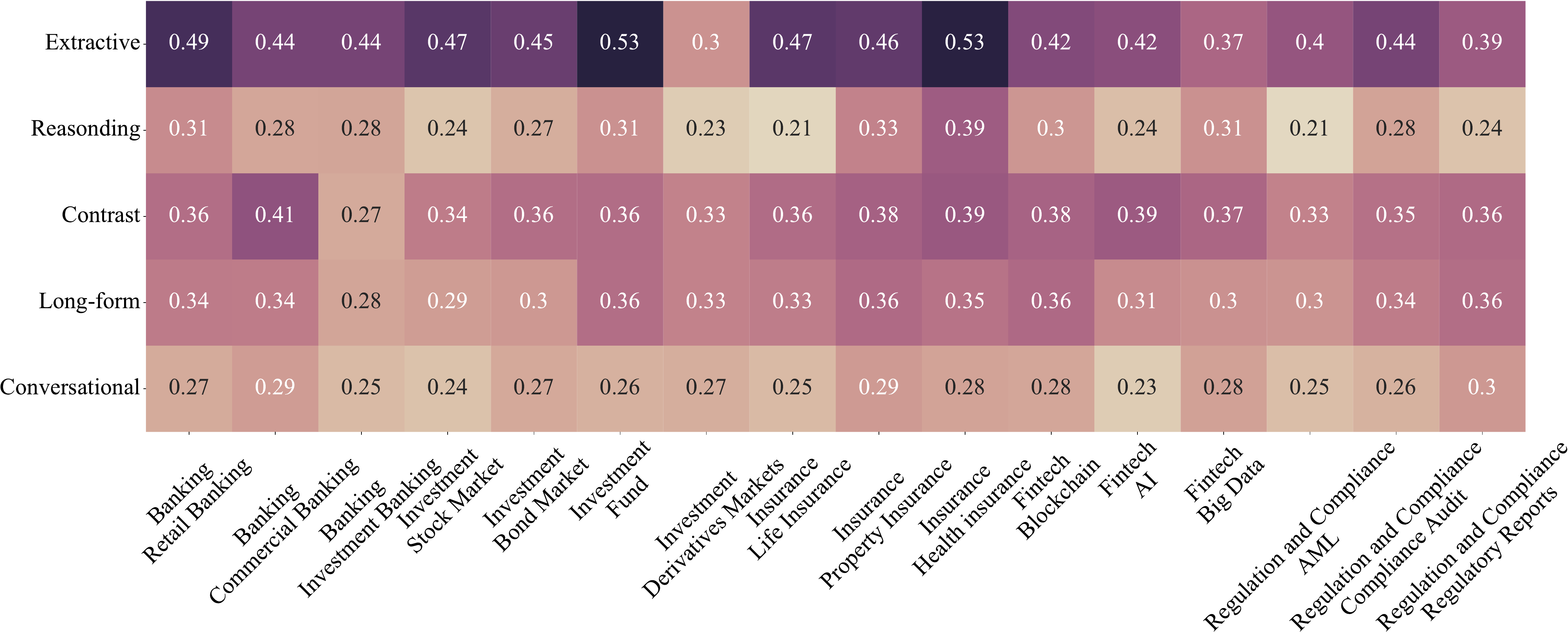}
    \caption{Rouge-L of matrix-based results of GTE-Qwen2-1.5B+Qwen2-72b on auto-generated subsets.}
    \label{fig:gte-qwen2-1.5b_TOP5-llama3-70b-instruct-auto}
\end{figure*}

\begin{figure*}[!h]
\centering
    \includegraphics[width=1\linewidth]{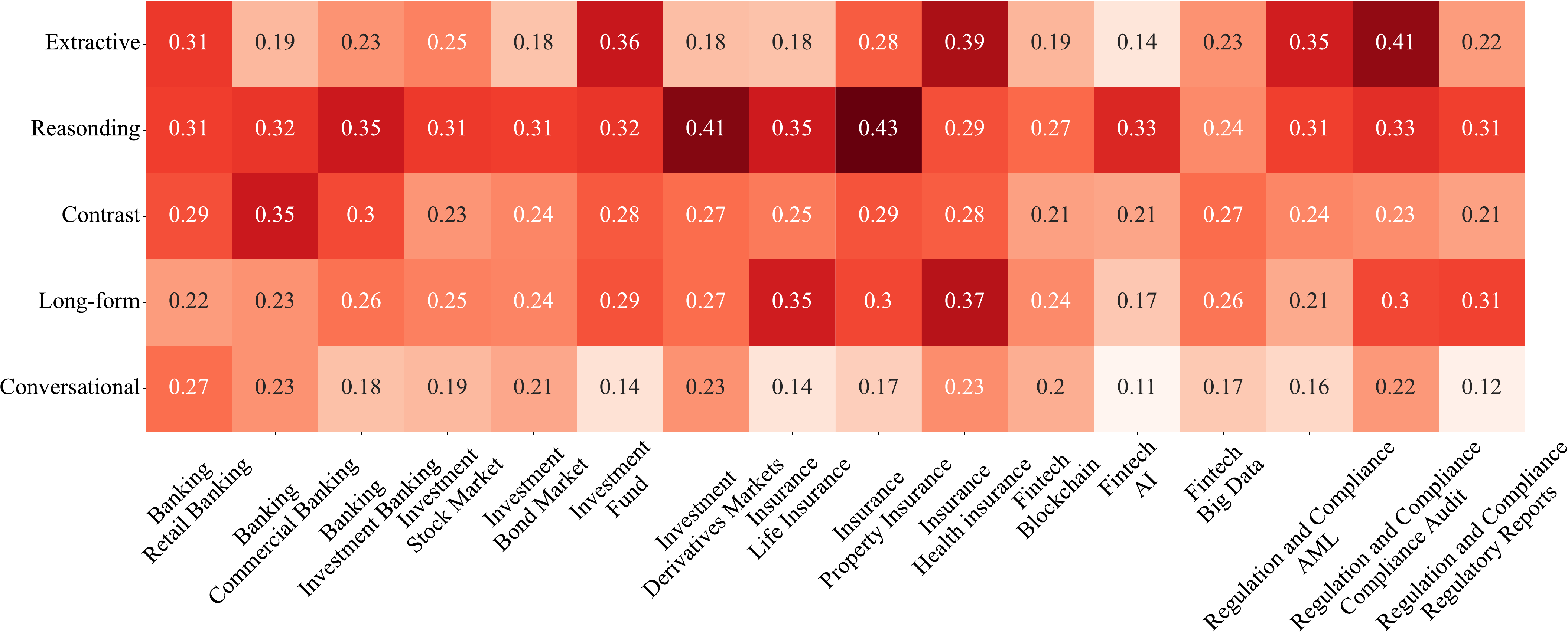}
    \caption{Rouge-L of matrix-based results of GTE-Qwen2-1.5B+Qwen2-72b on human-annotated subsets.}
    \label{fig:gte-qwen2-1.5b_TOP5-qwen-human}
\end{figure*}

\begin{figure*}[!h]
\centering
    \includegraphics[width=1\linewidth]{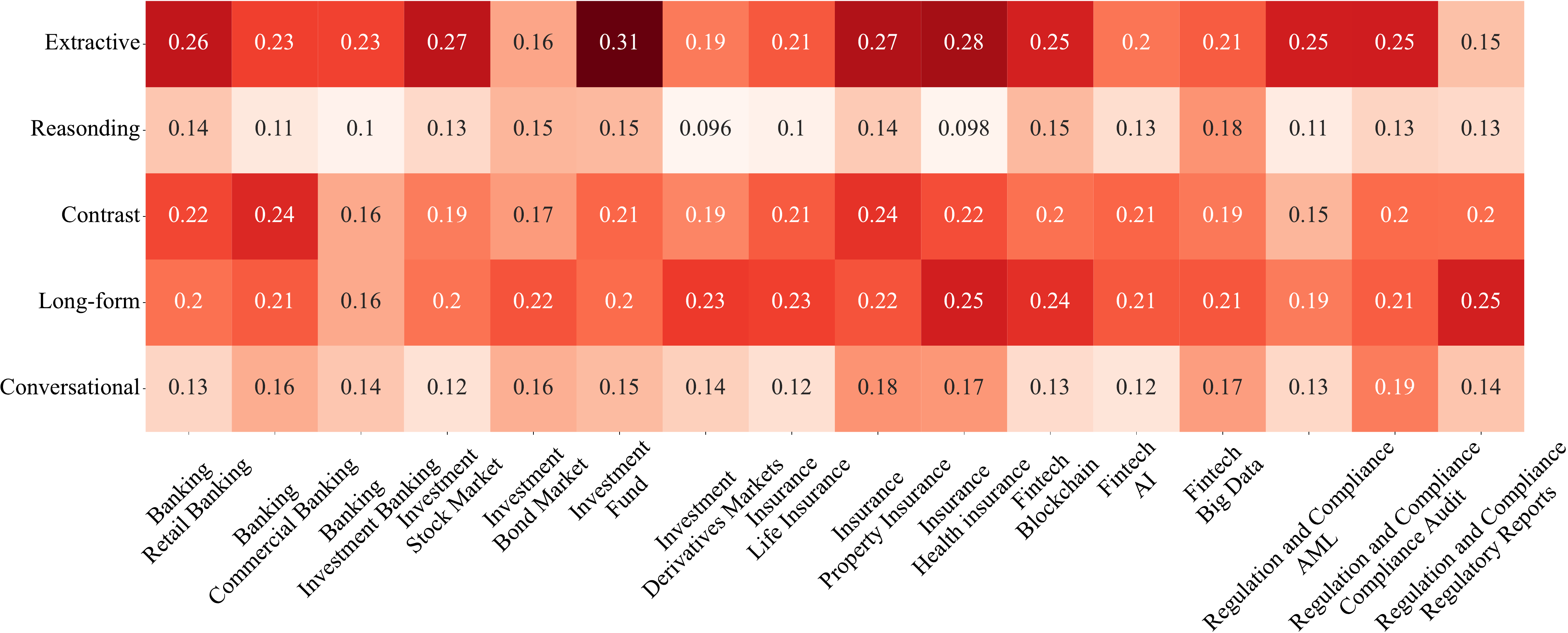}
    \caption{Rouge-L of matrix-based results of GTE-Qwen2-1.5B+Qwen2-72b on auto-generated subsets.}
    \label{fig:gte-qwen2-1.5b_TOP5-qwen-auto}
\end{figure*}

\begin{figure*}[!h]
\centering
    \includegraphics[width=1\linewidth]{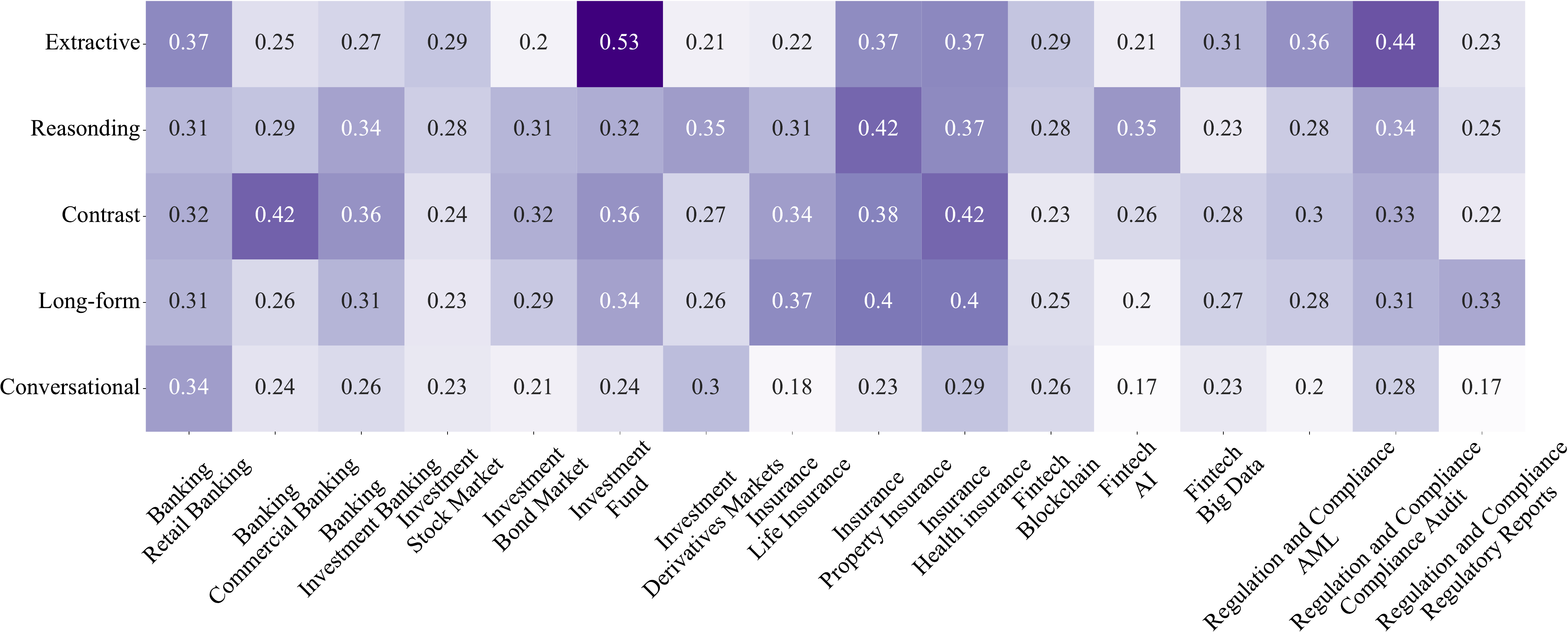}
    \caption{Rouge-L of matrix-based results of GTE-Qwen2-1.5B+deepseek-v2-chat on human-annotated subsets.}
    \label{fig:gte-qwen2-1.5b_TOP5-deepseek-human}
\end{figure*}

\begin{figure*}[!h]
\centering
    \includegraphics[width=1\linewidth]{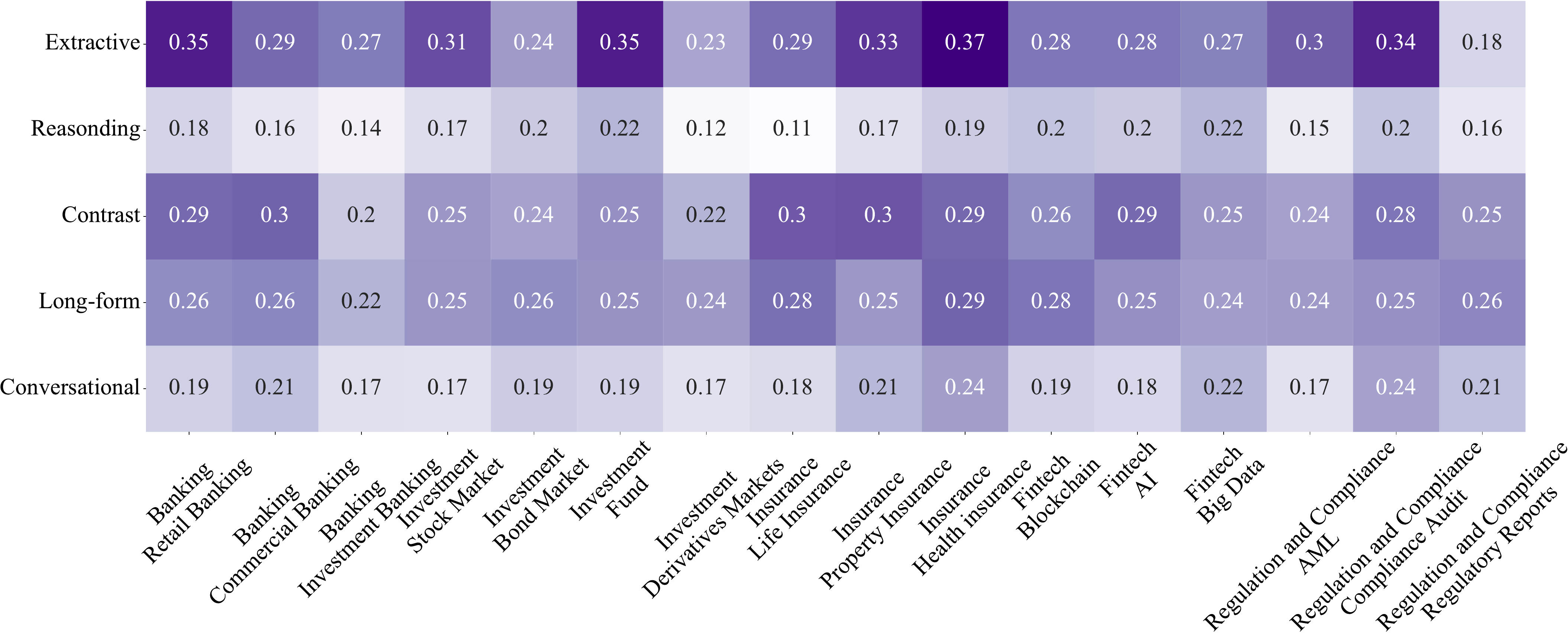}
    \caption{Rouge-L of matrix-based results of GTE-Qwen2-1.5B+deepseek-v2-chat on auto-generated subsets.}
    \label{fig:gte-qwen2-1.5b_TOP5-deepseek-auto}
\end{figure*}

\begin{figure*}[!h]
\centering
    \includegraphics[width=1\linewidth]{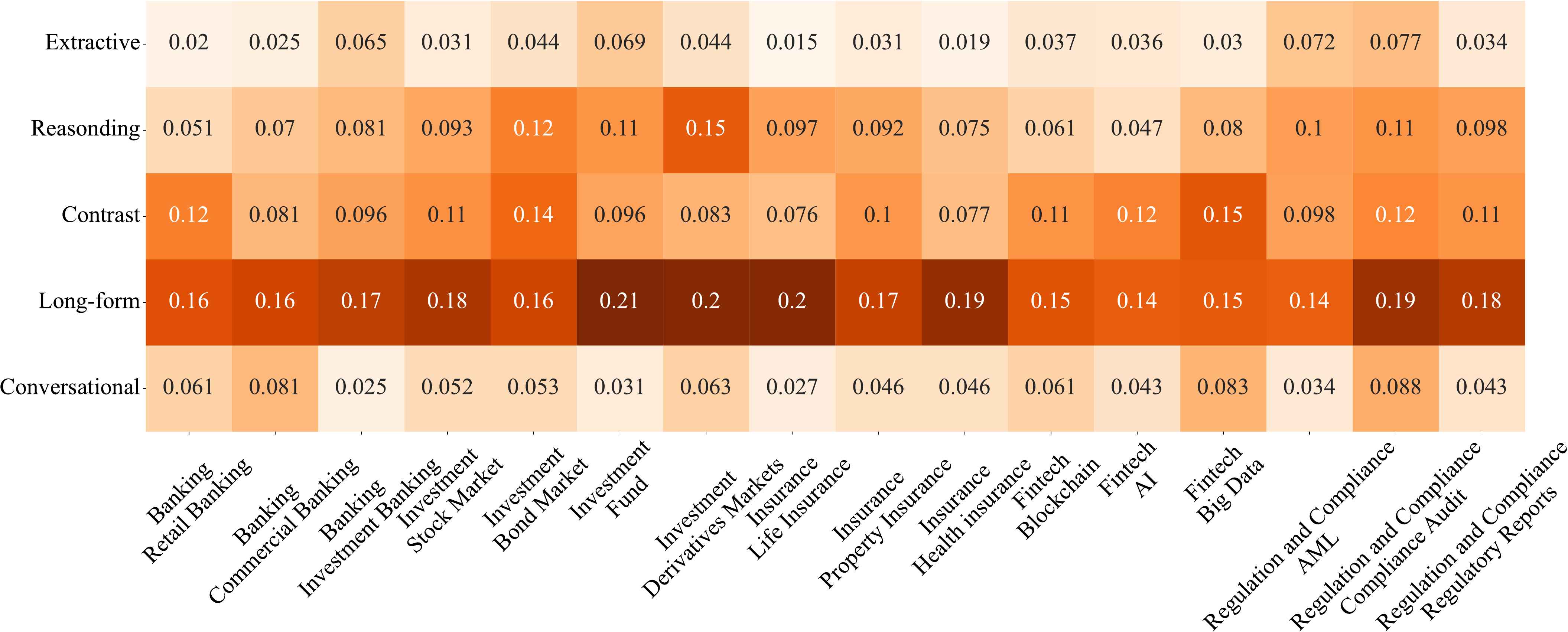}
    \caption{Rouge-L of matrix-based results of GTE-Qwen2-1.5B+Yi15-34B on human-annotated subsets.}
    \label{fig:gte-qwen2-1.5b_TOP5-yi-human}
\end{figure*}

\begin{figure*}[!h]
\centering
    \includegraphics[width=1\linewidth]{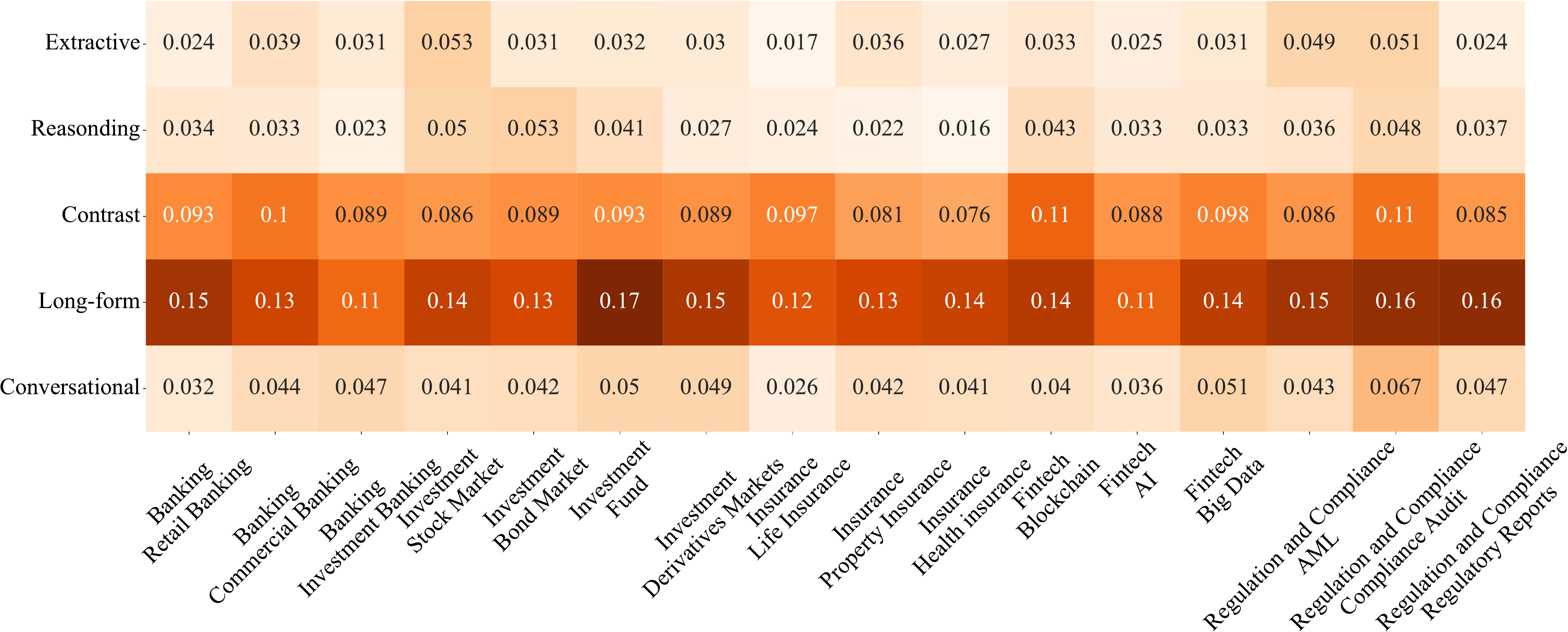}
    \caption{Rouge-L of matrix-based results of GTE-Qwen2-1.5B+Yi15-34B on auto-generated subsets.}
    \label{fig:gte-qwen2-1.5b_TOP5-yi-auto}
\end{figure*}

In this section, we present the supplementary matrix-based visualization results of our RAG models in Figures~\ref{fig:gte-qwen2-1.5b_TOP5-llama3-70b-instruct-auto},~\ref{fig:gte-qwen2-1.5b_TOP5-qwen-human},~\ref{fig:gte-qwen2-1.5b_TOP5-qwen-auto},~\ref{fig:gte-qwen2-1.5b_TOP5-deepseek-human},~\ref{fig:gte-qwen2-1.5b_TOP5-deepseek-auto},~\ref{fig:gte-qwen2-1.5b_TOP5-yi-human}, and~\ref{fig:gte-qwen2-1.5b_TOP5-yi-auto}.

\section{Human and GPT Instructions}\label{app:instructions}
\begin{figure*}
    \centering
    \includegraphics[width=1\linewidth]{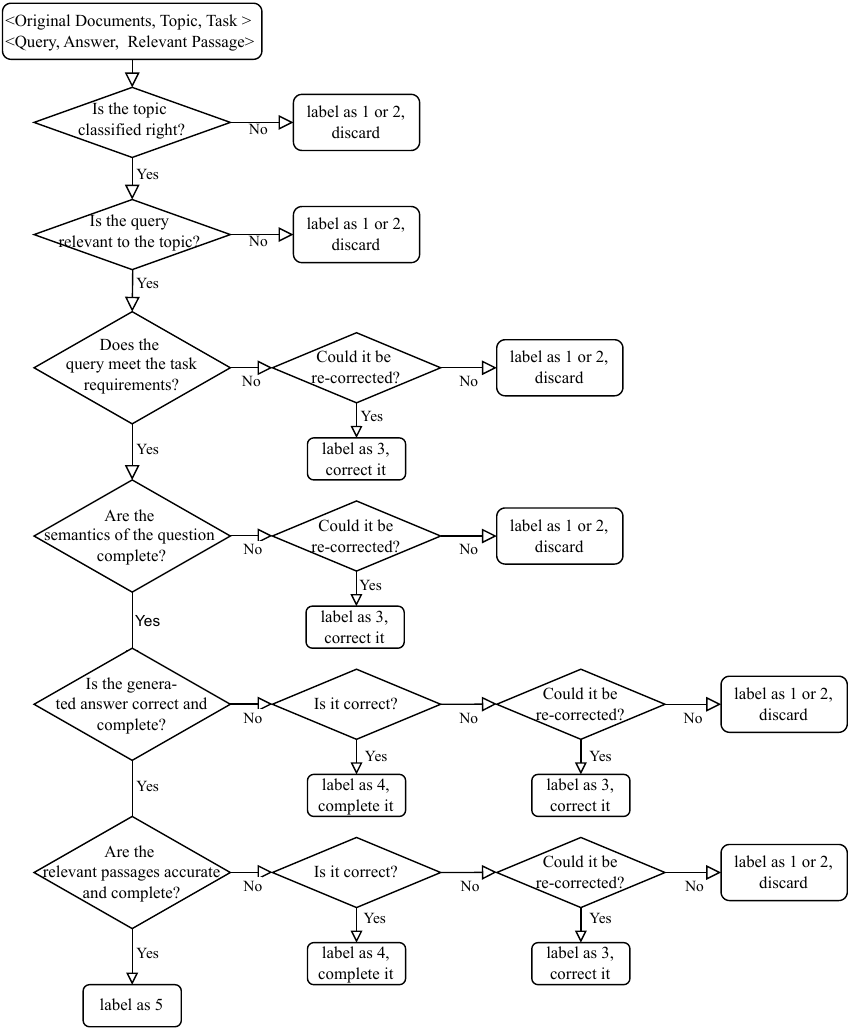}
    \caption{The pipeline of human annotation and correction for automatically generated data instances.}
    \label{fig:human-pipeline}
\end{figure*}

In this section, we provide detailed instructions we used for human annotation and GPT generation, including the topic-tree generation (Box~\ref{fig:topic-tree}), automated data generation (Boxs~\ref{box:gpt-data-generation},~\ref{box:gpt-data-quality-requirements-1}, and~\ref{box:gpt-data-quality-requirements-2}), automated data quality inspection (Box~\ref{box:gpt-data-filter}), and human annotation and correction (a flow chart, shown in Figure~\ref{fig:human-pipeline}). We also show detailed task requirements which support the GPT generation and human annotation in Tables~\ref{tab:task_requirements-1} and~\ref{tab:task_requirements-2}. 
% \balance




\begin{figure*}[h]
    \begin{tcolorbox}[title={Instructions for GPT-4 to generate a topic tree for the specific domain.}] 
    \#\# Background\newline 
You are a professional domain subcategory tree builder. I will provide you with the name of the root node for the domain type, and you should generate a comprehensive and diverse subcategory tree under that domain.\newline 
The output should be returned in JSON format. This JSON should include the following two properties:\newline 
- topic\_name: Represents the category name of the current tree node.\newline 
- sub\_topics: Represents the subcategory tree of the current tree node, which is a list of JSON data for that subcategory tree. If the current node is a leaf node (i.e., it has no subcategories), this property will be an empty list.\newline 
The data format requirements are as follows:\newline 
\{\newline 
    \hspace*{5mm}"topic\_name": The name of the category for this node,\newline 
    \hspace*{5mm}"sub\_topics": A list of JSON data for the subcategory tree under this node, with each item being JSON data of a subtree that also contains the "topic\_name" and "sub\_topics" properties.\newline 
\}\newline 
\#\# Name of the Root Node for the Domain Type\newline 
{domain\_name}
    \end{tcolorbox}
\caption{Instructions for GPT-4 to generate a topic tree for the specific domain.}
\label{box:topic-generation}
\end{figure*}

\begin{figure*}[h]
    \begin{tcolorbox}[title={Instructions for GPT-4 to classify the domain topic for the input document.}] 
    \#\# Background\newline
You are an intelligent document topic classification assistant. I am generating retrieval-augmented financial model multi-task evaluation data. This evaluation data is automatically generated by a large language model. I will provide the large language model with the following content: [financial subcategories of interest for the evaluation data, task description for the evaluation, documents in the knowledge base]. I need the large language model to generate: [user questions that align with the task description, corresponding correct answers, and document fragments that support those answers] based on the provided documents. I will provide you with a knowledge base document, and I need you to first classify whether the document falls within the scope of the financial domain, and if so, which topic subcategory it belongs to.\newline
\#\# Data Input Format\newline
The input consists of the following two parts:\newline
- Subcategory list: A list format of data, where each item in the list is JSON data representing a financial subcategory. This data includes the following attributes:\newline
    \hspace*{5mm}- id: An integer value representing the id of the financial topic subcategory. Your classification result should return only the subcategory id, not the subcategory name.\newline
    \hspace*{5mm}- topic\_name: A string representing the name of the financial topic subcategory.\newline
- Document content to be classified: A JSON formatted data, containing the following attributes:\newline
    \hspace*{5mm}- title: A string representing the document title.\newline
    \hspace*{5mm}- content: A string representing the document content.\newline
\#\# Generated Data Format\newline
You need to generate the value of the financial topic subcategory id that is most relevant to the document.\newline
If the document content is unrelated to finance, or does not relate to any provided financial topic subcategory, please return 0.\newline
Generate in JSON format, with the following data format:\newline
\{\newline
    \hspace*{5mm}"topic\_id": An integer value indicating the most relevant financial topic subcategory id for the document. If the document is unrelated to finance, please return 0.\newline
\}\newline
Note to generate only JSON formatted data, and do not generate any other characters.\newline
\#\# Subcategory List\newline
{topics\_str}\newline
\#\# Document Content to be Classified\newline
\{\newline
    \hspace*{5mm}"title": {title},\newline
    \hspace*{5mm}"content": {content},\newline
\}\newline
\#\# Most Relevant Subcategory ID for the Document
    \end{tcolorbox}
\caption{Instructions for GPT-4 to classify the domain topic for the input document.}
\label{box:gpt-classify-topic}
\end{figure*}

\begin{figure*}[h]
    \begin{tcolorbox}[title={Instructions for GPT-4 to automatically generate data instances.}] 
    \#\# Background\newline
You are an intelligent evaluation data generation assistant. I am generating retrieval-augmented financial model multi-task evaluation data. I require you to automatically generate evaluation data that is strongly relevant to the evaluation tasks. I will provide the following content: [financial topic subcategories of interest for the evaluation data, task descriptions and requirements, documents in the knowledge base]. I need you to generate evaluation data that is strongly relevant to the provided financial topic area and meets the evaluation task requirements. The evaluation data includes the following content:\newline
- User questions that align with the topic requirements and task descriptions\newline
- Corresponding correct answers\newline
- Document passages extracted from the original text that support those answers\newline
\#\# Quality Requirements for Data Generation\newline
    \hspace*{5mm}...(see details in Boxs~\ref{box:gpt-data-quality-requirements-1} and ~\ref{box:gpt-data-quality-requirements-2})\newline
\#\# Data Generation Process:\newline
1. First, determine if the document is a high-quality document. If the document is not closely relevant to the provided financial subtopic, has low informational content, is incomplete, has mixed formats, or does not meet the above requirements, then it is unsuitable for generating evaluation data. If the document is not suitable for generating domain-knowledge-related evaluation data, please return an empty list.\newline
2. If the document is high-quality, further assess whether it is suitable for generating relevant data for the provided evaluation task. If it is not suitable, please return an empty list.\newline
3. If the document is suitable for generating evaluation data relevant to the provided evaluation task and financial subtopic, please generate high-quality evaluation data.\newline
\#\# Generated Data Format Requirements\newline
The generated data should be returned in the form of a JSON data list, formatted as follows:\newline
[\newline
    \hspace*{5mm}\{\newline
        \hspace*{10mm}"thought\_process": A Chinese string representing your thought process while generating this data entry,\newline
        \hspace*{10mm}"question": A Chinese string representing the question posed by the user,\newline
        \hspace*{10mm}"answer": A list of strings representing all possible forms of the answer to that question,\newline
        \hspace*{10mm}"relevant\_passage": A list of Chinese strings representing relevant content excerpts from the original document that help answer the question. Please ensure the completeness of the extracted passages' information,\newline
    \hspace*{5mm}\},\newline
    \hspace*{5mm}...\newline
]\newline
\#\# Financial Subcategories of Interest for Evaluation Data\newline
\{topic\_name\}\newline
\#\# Task Description and Requirements\newline
\#\#\# Task Name\newline
\{task\_name\}\newline
\#\#\# Task Requirements\newline
\{task\_require\}\newline
\#\# Provided Document\newline
\{doc\_str\}\newline
\#\# List of Generated Data
    \end{tcolorbox}
\caption{Instructions for GPT-4 to automatically generate data instances.}
\label{box:gpt-data-generation}
\end{figure*}

\begin{figure*}[h]
    \begin{tcolorbox}[title={Quality requirements for data generation -- Part 1}] 
- Quality Requirements for Documents:\newline
    \hspace*{5mm}- First, determine whether the document is relevant to the domain being evaluated (financial subdomain). If it is not relevant, do not generate data.\newline
    \hspace*{5mm}- The content used to generate evaluation data should not involve any personal privacy of users, such as names, phone numbers, ID numbers, home addresses, etc. If the provided document contains private information, please return an empty list.\newline
    \hspace*{5mm}- The content used to generate evaluation data must be rigorous and of high quality; do not generate evaluation samples based on low-quality documents.\newline
    \hspace*{5mm}- If you believe the document is unsuitable for generating evaluation data for the provided task, please return an empty list.\newline
- Quality Requirements for Question Generation:\newline
    \hspace*{5mm}- User questions should be as realistic as possible, simulating what users genuinely care about when applying large language models for knowledge Q\&A in the financial domain.\newline
    \hspace*{5mm}- Questions must be semantically complete and unambiguous. The user's intent should be clear from the question content alone. Questions that rely on the content of the provided document to complete the context are strictly prohibited.\newline
    \hspace*{5mm}- Note that only when generating evaluation data for multi-turn dialogue capabilities should subsequent questions be ambiguous and dependent on previous dialogue content to clarify their semantics. In this case, subjects may be omitted or replaced with pronouns in later questions.\newline
    \hspace*{5mm}- Users do not provide documents when asking real questions; they only ask questions. Therefore, real user questions will not involve phrases like ``according to the given document...''. Such questions are strictly prohibited.\newline
    \hspace*{5mm}- The types of generated questions must strictly match the description of the evaluation task.\newline
    \hspace*{5mm}- The generated questions must be strongly relevant to the provided financial subtopic.\newline
    \hspace*{5mm}- Ensure the solvability of the generated questions. The answers in the generated data must be meaningful, and prohibited answers include ``none'', ``empty'', ``unable to answer based on the retrieved document'', etc.\newline
- Quality Requirements for Answer Generation:\newline
    \hspace*{5mm}- Only generate knowledge-rich data samples; the answers must contain substantial valuable information. Avoid generating vague or generic Q\&A pairs, especially answers like ``positive impact'', ``beneficial effect'', etc., which lack actual meaning.\newline
    \hspace*{5mm}- Answers must be consistent with the content of the provided document and should not contain factual inaccuracies or hallucinations.\newline
    \hspace*{5mm}- Ensure the accuracy and factual validity of the generated answers. The answers in the generated data must be meaningful; prohibited answers include ``none'', ``empty'', ``unable to answer based on the retrieved document'', etc.\newline
    \hspace*{5mm}- The format of answers can vary (\eg, numeric in Arabic or Chinese characters, various date formats), and please provide all possible forms of the answer in a string list format.\newline
\end{tcolorbox}
\caption{Quality requirements for data generation -- Part 1.}
\label{box:gpt-data-quality-requirements-1}
\end{figure*}

\begin{figure*}[h]
    \begin{tcolorbox}[title={Quality requirements for data generation -- Part 2}] 
- Quality Requirements for Relevant Passage Extraction:\newline
    \hspace*{5mm}- Must accurately provide document passages that support the answer; these passages must come from the original text of the provided document and cannot be altered.\newline
    \hspace*{5mm}- The extracted relevant passage content must be complete and coherent, without missing contextual meaning.\newline
- Overall Quality Requirements for Generated Evaluation Samples:\newline
    \hspace*{5mm}- Please strictly follow the evaluation task requirements to generate evaluation data that corresponds to that task's capabilities; for instance, multi-hop reasoning tasks must generate questions that require multiple inferences from the retrieved documents to answer, rather than being answerable in a single reading.\newline
    \hspace*{5mm}- The question-answer pairs generated must be answerable based on the content of the document, meaning understanding the document content is crucial to answering the question, and the role of the reference document cannot be ignored in the dialogue.\newline
    \hspace*{5mm}- Multiple high-quality evaluation data entries can be generated, but the high quality of the generated data must be guaranteed.\newline
    \hspace*{5mm}- Ensure precision in generated data rather than recall; only generate data that fully meets requirements, prohibiting data with low confidence.\newline
    \hspace*{5mm}- Generated data must meet task requirements and be strongly relevant to the target task and financial domain. If the document cannot generate any task-related data, please return an empty list.\newline
    \hspace*{5mm}- Ensure diversity in the generated data; do not generate multiple identical or closely similar evaluation data entries.
\end{tcolorbox}
\caption{Quality requirements for data generation -- Part 2.}
\label{box:gpt-data-quality-requirements-2}
\end{figure*}

\begin{figure*}[h]
    \begin{tcolorbox}[title={Instructions for GPT-4 to inspect the quality of the generated instance -- Part 1}] 
    \#\# Background\newline
You are a professional data quality evaluator and corrector. I will provide you with evaluation data generated by a large language model (related to the financial domain), and your task is to assess the quality of this generated data and make corrections when necessary. The quality of the generated data is classified into three levels:\newline
- 0: The quality of the generated data is very poor, and it cannot be suitably corrected to become high-quality data.\newline
- 1: The quality of the generated data is average; the generated questions, answers, or extracted relevant passages do not meet the requirements, but they can be corrected to become high-quality data.\newline
- 2: The quality of the generated data is very high and does not require correction.\newline
\#\# Background Knowledge -- Data Generation Process:\newline
    \hspace*{5mm}...(summarization of data generation process)\newline 
\#\# Input Content for Data Quality Evaluation Task:\newline
1. A long document in the financial domain used for generating data.\newline
2. The financial subtopic that the generated data should conform to.\newline
3. The description and requirements of the evaluation subtask to which the generated data belongs.\newline
4. The evaluation data generated by the large language model is to be assessed. The format of this data is a JSON list containing:\newline
[\newline
\hspace*{5mm}\{\newline
        \hspace*{10mm}"thought\_process": A Chinese string representing the thought process of the large language model when generating this data entry.\newline
        \hspace*{10mm}"question": A Chinese string representing the question posed by the user,\newline
        \hspace*{10mm}"answer": A list of strings representing all possible forms of the answer to that question.\newline
        \hspace*{10mm}"relevant\_passage": A list of Chinese strings representing relevant content excerpts from the original document that help answer the question. Please ensure the completeness of the extracted passages' information.\newline
    \hspace*{5mm}\},\newline
    \hspace*{5mm}...\newline
]
    \end{tcolorbox}
\caption{Instructions for GPT-4 to inspect the quality of the generated instance -- Part 1.}
\label{box:gpt-data-filter}
\end{figure*}

\begin{figure*}[h]
    \begin{tcolorbox}[title={Instructions for GPT-4 to inspect the quality of the generated instance -- Part 2}] 
\#\# Data Quality Evaluation Requirements\newline
1. Determine whether the generated questions are related to the provided financial subtopic.\newline
2. Assess whether the generated questions meet the requirements of the evaluation subtask, paying particular attention to whether questions for multi-hop reasoning tasks require multi-hop reasoning.\newline
3. Check if the answers to the generated questions are correct and whether they can be fully answered based on the provided long document.\newline
4. Evaluate whether the extracted relevant passages from the original text are complete and sufficiently support the full answer to the generated questions.\newline
\#\# Output Requirements and Format for Evaluation and Correction Results\newline
Only when you assess the quality of the data as 1 should you make corrections; no corrections are needed for 0 or 2.\newline
During the data quality evaluation process, pay special attention to the following key points:\newline
\hspace*{5mm}- For questions of the form ``yes or no'' where the answer is usually ``yes'' or similar affirmative responses, please mark the quality as 0. This is because it is generally impossible to generate data pairs with a ``no'' answer, and such generated data would bias our dataset; therefore, please remove this type of generated data.\newline
\hspace*{5mm}- For multi-hop reasoning questions, pay special attention to whether the question requires multi-hop reasoning, meaning the (retrieval-augmented) large language model needs to engage in at least two steps of ``thinking-answering'' reasoning to fully resolve the issue. If the question only adds complex conditions but can still be solved with a single inference, the quality of such generated data should be marked as 0 or 1. If it can be corrected based on the original document, mark it as 1 and correct it. If it cannot be corrected, mark it as 0.\newline
The evaluation results should be returned in JSON format, with the specific format and requirements as follows:\newline
\{\newline
    \hspace*{5mm}"evaluation": An integer value indicating the assessment result of the generated data quality, with values in [0, 1, 2].\newline
    \hspace*{5mm}"corrected\_result": A JSON list format of the corrected results for data assessed as quality 1, making them high-quality evaluation data. If the evaluation quality is 0 or 2, this attribute should be None. Note: The data format and types should be completely consistent with the input evaluation data generated by the large language model; only the contents of the internal attributes are corrected.\newline
\}\newline
\#\# Long Document in the Financial Domain Used for Data Generation\newline
\{doc\_str\}\newline
\#\# Financial Subtopic that the Generated Data Should Conform to\newline
\{topic\_name\}
\#\# Description and Requirements of the Evaluation Task to Which the Generated Data Belongs\newline
\#\#\# Task Name\newline
\{task\_name\} \newline
\#\#\# Task Requirements\newline
\{task\_require\}\newline
\#\# Evaluation Data Generated by the Large Language Model\newline
\{gen\_datas\}\newline
\#\# Evaluation and Correction Results\newline
    \end{tcolorbox}
\caption{Instructions for GPT-4 to inspect the quality of the generated instance -- Part 2.}
\label{box:gpt-data-filter}
\end{figure*}

\begin{table*}[]
    \centering
    \begin{tabular}{p{2cm}p{12cm}}
    \toprule
        Task & Requirement  \\
    \midrule
         Extractive QA & This task is designed to evaluate the ability of retrieving enhanced financial large language models to answer one-hop questions. That is, the user's question does not need to do multi-hop thinking, and the answer to the question can be directly found in the search document and extracted as an answer. \newline
- Please note the distinction between this task and multi-hop inference problems.\\
    \midrule
    Multi-hop Reasoning & This task aims to evaluate the ability of a retrieve-enhanced financial grand language model to answer questions involving multi-hop reasoning. That is, the answer cannot be found directly in the external document retrieved, and **the model needs to do at least two hops of reasoning** to arrive at the final answer according to the external information provided by the document or its own knowledge.\newline
- Do not generate questions that can be answered with one-hop reasoning.\newline
- Evaluation data generation to evaluate multi-hop inference capability mainly includes the following two categories:\newline
    \hspace*{5mm}1. First identify the ``entity-relationship'' link composed of multiple entities with information progressive relationship in the document, and then generate multi-hop inference data according to the relationship link. That is, there should be at least two unknown information points in the proposed question (**and the unknown information in the middle node is necessary for solving the final question**). To solve the final answer, the LLM to be evaluated needs to perform information retrieval and reasoning on the previously unknown information points to obtain the dependency information for solving the final answer, and then solve the final answer. Trying to satisfy the content of the question is a more obvious need for multi-hop reasoning.\newline
    \hspace*{5mm}2. If you need to perform financial calculations based on the information provided in the document, ensure that the questions and answers are accurate.\newline
- If I provide one piece of document data, generate the second type of multi-hop inference data, which is the problem that requires financial calculation based on the information provided in the document.\newline
- If I provide multiple document data, generate the first type of multi-hop inference data. That is to identify the ``entity-relationship'' link composed of multiple entities with information transfer relationship in the document, and ensure that the ``entity-relationship'' link is through all the provided documents, and then generate multi-hop inference data according to the relationship link. Please ensure that the generated multi-hop inference problem cannot be solved by only one document content, ensure that all documents provided are valuable for solving the generated inference problem.\newline
- Be careful not to directly write out the complete content of each step of information transmission in the question, especially do not say that the middle answer is written in the question, otherwise the multi-hop reasoning problem will degenerate into a one-hop reasoning problem. \\
    \bottomrule
    \end{tabular}
    \caption{Requirements of tasks for human and GPT generation -- Part 1.}
    \label{tab:task_requirements-1}
\end{table*}

\begin{table*}[]
    \centering
    \begin{tabular}{p{2cm}p{12cm}}
    \toprule
        Task & Requirement  \\
    \midrule
         Contrast QA & This task is designed to evaluate the ability of a retrieve-enhanced financial large language model to answer questions involving contrast classes. That is, the question involves comparing two aspects of the transaction, and the corresponding answer needs to provide a correct and comprehensive comparison and summary of results.\newline
- When I provide multiple document data, please ensure that the generated question-answer data is cross-document, \ie, the need to answer the question requires the help of all the provided document data. Based on only one or a few of them can lead to incomplete answers.\\
    \midrule
    Long-form QA & This task is designed to evaluate the ability to retrieve enhanced financial large language models when answering questions with longer answers. Such as introducing classes and summarizing class problems.\newline
- Ensure that the answers to the generated data are comprehensive enough to cover all aspects of the user's questions.\newline
- When I provide multiple document data, please ensure that the generated question-answer data is cross-document, \ie, the need to answer the question requires the help of all the provided document data. Based on only one or a few of them can lead to incomplete answers.  \\
    \midrule
    Conversation QA & This task is designed to evaluate the ability to retrieve enhanced financial large language models to do multiple rounds of conversations. That is, the generated data should be in the form of multiple rounds of conversations.\newline
- Therefore, the document is required to be rich enough in contextual information to support the generation of multiple rounds of conversations.\newline
- Take care to ensure the dependency between the generated multiple rounds of dialogue, especially the dependency of the content of the question, that is, the subject of the question in the second and later rounds is missing, or is a pronoun, resulting in ambiguous semantics. Understanding the full intent of subsequent rounds of questions requires a full understanding of what was said in previous rounds.\newline
- The generated data should be stored as a JSON list for multiple rounds of Q\&A information.\newline
- I may provide multiple document data, in this case, please ensure that the generated multi-round conversation data is cross-document and able to use all the content of the provided document. \\
    \bottomrule
    \end{tabular}
    \caption{Requirements of tasks for human and GPT generation -- Part 2.}
    \label{tab:task_requirements-2}
\end{table*}

\end{document}